\documentclass[11pt]{article}
\usepackage{acl}

\usepackage{times}
\usepackage{latexsym}
\usepackage[T1]{fontenc}
\usepackage[utf8]{inputenc}
\usepackage{microtype}
\usepackage{inconsolata}
\usepackage{graphicx}
\usepackage{xspace}
\usepackage{amsmath}
\usepackage{amssymb}
\usepackage{algorithm}
\usepackage{algorithmic}
\usepackage{enumitem}
\usepackage{multirow}
\usepackage{adjustbox}
\usepackage{booktabs}
\usepackage{xcolor}
\usepackage{colortbl}
\usepackage{nicematrix}
\usepackage{subfigure}

\newcommand{\m}{ChainUQ\xspace}
\title{\m: Reasoning Consistency-Aware Uncertainty Quantification for Large Language Models}

\author{
Dahai Yu,
Rongchao Xu,
Lin Jiang,
Ximiao Li,
Guang Wang
\\
Florida State University, Tallahassee, Florida, USA
\\
\small{
\texttt{\{dahai.yu, rxu, lin.jiang, xl24g\}@fsu.edu, guang@cs.fsu.edu}
}
}

\begin{document}
\maketitle

\begin{abstract}
While large language models (LLMs) exhibit impressive reasoning capabilities, response-level confidence may remain unreliable when intermediate claims conflict with the final conclusion. Therefore, effective uncertainty quantification (UQ) is required to capture logical inconsistencies within the reasoning chain, not just the correctness of the final output. 
Current approaches have two major limitations: (1) their reliance on token-level probabilities fails to capture reasoning consistency, and (2) they lack mechanisms to dynamically calibrate confidence using the structural logic of the generated chain. 
To advance existing research, we introduce \m, a reasoning consistency-aware uncertainty quantification framework for LLMs. \m consists of two key technical components: an alignment-aware lightweight UQ module that estimates a raw intrinsic model confidence score from frozen features aligned to the final conclusion, and a reasoning consistency-aware calibrator that refines this score using reasoning-chain consistency evidence. 
Evaluations across diverse in-distribution and out-of-distribution benchmarks show that \m consistently improves response-level uncertainty estimation, achieving an average 3.1\% relative gain in AUROC and up to 45.0\% relative reduction in ECE, 
and can be directly transferred to new settings without additional fine-tuning.
\end{abstract}
\section{Introduction}\label{sec:introduction}

Chain-of-thought (CoT) prompting enables multi-step reasoning in large language models (LLMs), yet managing their reliability remains a critical challenge \citep{wei-etal-2022-chain-of-thought}. An LLM can produce a reasoning chain that appears coherent but still leads to an incorrect conclusion, as one or more intermediate claims may be unsupported, internally inconsistent, or misaligned with the final answer. Such reasoning errors can remain difficult to detect when the response appears fluent, and may still undermine response reliability even if the final answer happens to be correct. Therefore, response-level uncertainty quantification (UQ) should go beyond final-answer correctness and assess the reliability of the underlying reasoning chain.

In this work, we aim to develop a response-level UQ framework for LLMs that explicitly leverages reasoning consistency signals within the reasoning chain, which is capable of assessing whether individual reasoning steps are logically supported and whether the overall chain sufficiently justifies the final answer. However, current approaches face two key challenges. \textbf{First}, they lack an effective mechanism to calibrate confidence using the logical structure of the reasoning chain. A response may appear fluent and confident even when its intermediate steps are weakly supported or internally inconsistent. When the reasoning trace is evaluated as a whole, the resulting confidence score can obscure these step-level logical failures, making them difficult to detect~\citep{lin-etal-2022-truthfulqa}. \textbf{Second}, they struggle to balance efficiency with reasoning-structure modeling, i.e., the ability to model logical relations among reasoning steps. Black-box self-consistency methods~\citep{manakul-etal-2023-selfcheckgpt} can capture semantic discrepancies but require multiple sampled responses, whereas efficient probability- or representation-based methods~\citep{azaria-mitchell-2023-internal-state,yona-etal-2024-intrinsic-uncertainty} rely on local or global features without explicitly incorporating reasoning consistency information across intermediate steps.

To address these challenges, we introduce \m, a reasoning consistency-aware UQ framework for LLMs. Given a single response generated by a frozen white-box LLM, \m estimates response-level confidence by evaluating both the final answer and the underlying reasoning chain. 
It produces a calibrated confidence score without requiring repeated generation or retraining the base LLM. Rather than generating new reasoning paths, \m performs post-generation evaluation through two complementary signals: \textit{intrinsic model confidence}, which reflects the model's internal confidence in the final conclusion, and \textit{reasoning-chain consistency evidence}, which captures whether the intermediate reasoning steps logically support the conclusion. Specifically, \m contains two key components: an \textbf{alignment-aware lightweight UQ} module that estimates intrinsic model confidence from conclusion-aligned hidden representations of the frozen LLM, and a \textbf{reasoning consistency-aware calibration} module that uses reasoning-chain consistency evidence to calibrate this confidence score. Through these two novel designs, \m can better identify unreliable responses that appear confident, thereby improving uncertainty discrimination and calibration without retraining the backbone LLM or repeated generation.

\begin{itemize}
    \item \textbf{Conceptually,} we extend response-level UQ beyond final-answer uncertainty by modeling both intrinsic model confidence and reasoning-chain consistency evidence. By explicitly grounding uncertainty in step-level logical dependencies, reasoning consistency-aware UQ helps identify reasoning flaws that may not be reflected by surface-level fluency or final-answer confidence.

    \item \textbf{Technically,} we introduce \m, a single-response reasoning consistency-aware UQ framework for LLMs with two novel complementary modules. An alignment-aware lightweight UQ module estimates intrinsic model confidence from conclusion-aligned internal features, and a reasoning consistency-aware calibration module refines this confidence using reasoning-chain consistency evidence. This design requires neither backbone LLM retraining nor repeated generation.

    \item \textbf{Empirically,} we evaluate \m using four datasets in both in-distribution and out-of-distribution scenarios. Extensive experiments show that \m consistently outperforms state-of-the-art baselines, achieving an average 3.1\% relative gain in AUROC and up to 45.0\% relative reduction in ECE while demonstrating robust transferability. 
\end{itemize}
\section{Preliminary}\label{sec:preliminary}

Given question-context inputs $x=(q,\mathcal{D})$, where $q$ is the question text and $\mathcal{D}$ is the provided context, we consider a frozen white-box LLM $M$, where white-box means that its internal states and generated reasoning chain are accessible for post-generation analysis. Our goal is to quantify the trustworthiness of the generated response with a calibrated confidence score, rather than selecting the best answer from multiple sampled candidates.

\subsection{Reasoning Chain Decomposition}\label{sec:decomposition}

An LLM model produces one reasoning chain $r$, which can be decomposed into $K$ intermediate reasoning claims and a final conclusion claim:
\begin{equation}
\mathcal{C}(r)=\{c^{r}_1,\ldots,c^{r}_K,c^{c}\}.
\end{equation}
Here $\mathcal{C}(r)$ is the decomposed claim set, $c_k^{r}$ is the $k$-th intermediate reasoning claim, and $c^c$ is the final conclusion claim.
This decomposition serves two purposes. First, it converts reasoning chains with different textual formats into a unified claim-level representation, making the analysis less dependent on the prompting style. Second, it separates the final conclusion from the intermediate reasoning claims. The final conclusion is used to center confidence estimation on the final answer, while the intermediate claims provide reasoning-chain consistency evidence.

\subsection{Consistency-Aware Labels}\label{sec:consistency-aware}
Based on the decomposition strategy in Section~\ref{sec:decomposition}, we assign different labels to the final conclusion and the intermediate reasoning claims. The conclusion claim $c^{c}$ receives a final-conclusion correctness label $y \in \{0,1\}$, indicating whether the final answer is factually correct under the given question and context. For the intermediate reasoning claims, we assign step-level consistency labels $y^{r}_1,\ldots,y^{r}_K \in \{0,1\}$ to characterize their reliability within the reasoning chain. Unlike the final-conclusion correctness label, these labels focus on the reliability of intermediate reasoning rather than the conclusion itself. Specifically, they assess whether each claim is factually supported by the provided context, logically coherent with preceding steps, and useful for supporting the final conclusion. We obtain them automatically using a stronger LLM as the judge.

\begin{figure*}[t]
\centering
\includegraphics[width=\textwidth]{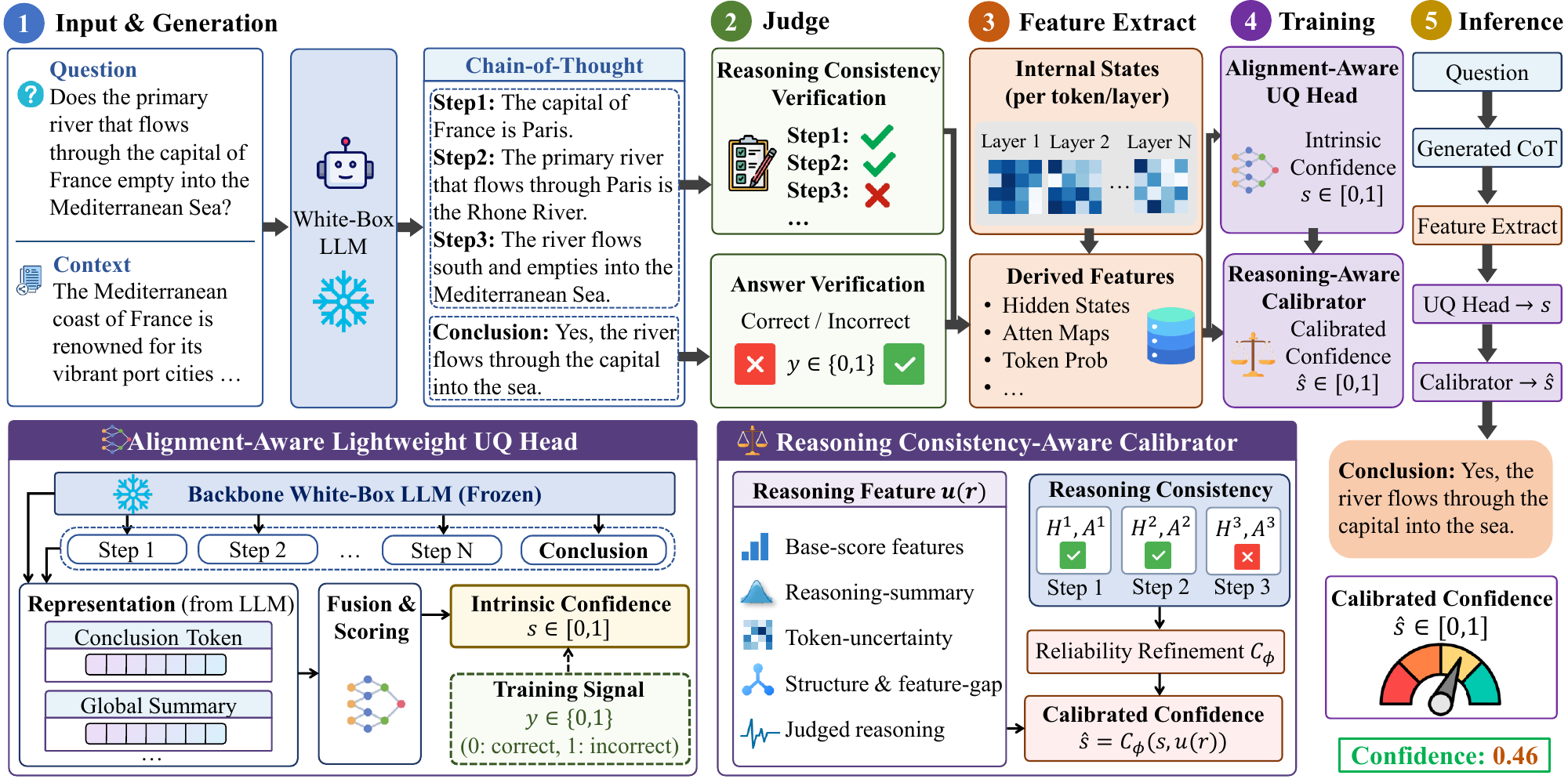}
\caption{Overview of \m. The framework follows a five-stage pipeline: a frozen white-box LLM generates a CoT trace; a judge verifies answer correctness and reasoning consistency; internal feature extraction; a lightweight UQ head predicts a raw confidence score $\boldsymbol{s}$, which is then refined by a reasoning consistency-aware calibrator into a calibrated confidence score $\hat{\boldsymbol{s}}$; finally, the model returns the conclusion together with its calibrated confidence.}
\label{fig:framework}
\end{figure*}

\section{Methodology}\label{sec:methodology}
In this section, we present the core methodology of~\m. 
As illustrated in Figure~\ref{fig:framework}, \m is a post-generation reasoning consistency-aware UQ framework for question-answering (QA) tasks. Given a QA input, \m outputs a conclusion as the answer and a calibrated confidence score indicating its trustworthiness. 
In~\m, the question and optional context are fed into a frozen white-box LLM to generate the full reasoning chain. 
After judge and feature extraction, the resulting internal states and derived features are processed by two sequential UQ modules. 
The first module estimates the intrinsic model confidence in the conclusion, while the second module calibrates this estimate using reasoning-chain consistency evidence.

Section~\ref{sec:two_stage_uq} first presents the response-level UQ formulation. Section~\ref{sec:lightweightUQ} then introduces the alignment-aware lightweight UQ module, and Section~\ref{sec:calibration} describes the reasoning consistency-aware calibration module. 
A more detailed pipeline description is provided in Appendix~\ref{sec:app_pipeline}.

\subsection{Response-Level Reasoning Consistency-Aware UQ Formulation}\label{sec:two_stage_uq}
We consider a frozen white-box LLM \(M\) prompted by \(x=(q,\mathcal{D})\), where \(q\) denotes the question and \(\mathcal{D}\) is an optional context.  The model generates a reasoning chain \(r\), which consists of multiple reasoning claims and one final conclusion.
After decomposing \(r\) into claims, we tokenize each claim with the LLM tokenizer and collect the corresponding frozen hidden states of all tokens as \(\mathbf{F}=[f_1,\ldots,f_T]\in\mathbb{R}^{T\times d}\), where \(f_t\in\mathbb{R}^{d}\) denotes the hidden state of the \(t\)-th token, \(T\) is the total number of tokens, and \(d\) is the hidden dimension. 
Following Section~\ref{sec:decomposition}, let $m^{c}\in\{0,1\}^{T}$ denote the binary mask over the contiguous token span that realizes the conclusion claim $c^{c}$ in the generated response.
In \m, we formulate the response-level UQ problem in two stages:
\begin{equation} \label{eq:2}
\boldsymbol{s} = F_{\theta}(\mathbf{F},m^{c}), \qquad \hat{\boldsymbol{s}}=C_{\phi}(\boldsymbol{s},u(r)).
\end{equation}
In the first stage \(F_{\theta}\) denotes the alignment-aware lightweight UQ estimator, which estimates the intrinsic model confidence \(\boldsymbol{s}\) as the raw confidence from the token hidden states \(\mathbf{F}\) and the binary conclusion mask \(m^c\). In the second stage, given the reasoning-chain consistency evidence \(u(r)\in\mathbb{R}^{d_u}\), the reasoning consistency-aware calibrator \(C_{\phi}\) further refines \(\boldsymbol{s}\) into the final calibrated confidence score \(\hat{\boldsymbol{s}}\).
This two-stage design has two main advantages. 
First, it separately models conclusion confidence and reasoning-chain reliability, reducing their mutual interference during trace evaluation. 
Second, it reuses the internal states and reasoning-chain evidence obtained from one generated reasoning chain, without repeated sampling or backbone LLM fine-tuning.

Ideally, the calibrated confidence \(\hat{\boldsymbol{s}}\) should approximate the probability that the generated conclusion is correct:
\begin{equation} \label{eq:3}
\hat{\boldsymbol{s}}\approx \mathbb{P}(y=1\mid \boldsymbol{s},u(r)),
\end{equation}
where \(y\in\{0,1\}\) is the response-level correctness label.

\subsection{Alignment-Aware Lightweight UQ}\label{sec:lightweightUQ}
After obtaining the internal states and derived features of the reasoning chain \(r\), the alignment-aware lightweight UQ module estimates the model's raw confidence in the answer, i.e., the intrinsic model confidence \(\boldsymbol{s}\). 

\textbf{Step 1: Token-state projection.}
Given the token hidden states \(\mathbf{F}=[f_1,\ldots,f_T]\) defined in Section~\ref{sec:two_stage_uq}, we first apply layer normalization \(\mathrm{LN}(\cdot)\) and a learned projection \(\Phi(\cdot)\) to each \(f_t\):
\begin{equation}
\mathbf{H}=
\big[
\Phi(\mathrm{LN}(f_1)),\ldots,\Phi(\mathrm{LN}(f_T))
\big].
\end{equation}
Here, \(\mathbf{H}\in\mathbb{R}^{T\times d_h}\) denotes the projected token hidden states, where each row corresponds to one token in the reasoning chain and \(d_h\) is the projected hidden dimension.

\textbf{Step 2: Response-level context aggregation.}
We then construct a global sample state \(h^{\,g}\) from the projected token states \(\mathbf{H}\) using gated mean/max pooling:
\begin{equation} \label{eq:5}
\begin{aligned}
h^{\,g} &= \mathrm{GPool}\!\big(
\mathrm{MeanPool}(\mathbf{H},a), \\
&\qquad\qquad\ \mathrm{MaxPool}(\mathbf{H},a)
\big).
\end{aligned}
\end{equation}
Here, \(a\in\{0,1\}^{T}\) is the valid-token attention mask, where \(a_t=1\) indicates that the \(t\)-th token is valid. 
\(\mathrm{MeanPool}(\mathbf{H},a)\) and \(\mathrm{MaxPool}(\mathbf{H},a)\) aggregate over the valid tokens selected by \(a\), and \(\mathrm{GPool}\) denotes a learned gate followed by a residual MLP. 
The resulting \(h^{\,g}\) serves as a coarse response-level anchor for later comparison with the conclusion representation, rather than directly estimating confidence alone.

\textbf{Step 3: Conclusion-aware state encoding.}
To explicitly mark the final-conclusion tokens, we add a learned marker embedding \(E(m^c)\) to the projected token states \(\mathbf{H}\), where \(E(\cdot)\) maps the conclusion mask \(m^c\) into token-wise marker embeddings.
A lightweight scope encoder \(\operatorname{Enc}_{\mathrm{sc}}\) then produces the scope-aware states:
\begin{equation}
\mathbf{H}^{\mathrm{sc}}=\operatorname{Enc}_{\mathrm{sc}}(\mathbf{H}+E(m^{c}),a),
\end{equation}
where \(\mathbf{H}^{\mathrm{sc}}\in\mathbb{R}^{T\times d_h}\). 
In this way, the token representations are updated with explicit conclusion-span information while still attending to the valid reasoning context.

From \(\mathbf{H}^{\mathrm{sc}}\), we extract a scoped global representation and two conclusion-centered representations:
\begin{equation}\label{eq:7}
\begin{aligned}
\bar{h}^{\,g}_{\mathrm{sc}}&=\mathrm{MeanPool}(\mathbf{H}^{\mathrm{sc}},a), \\
\bar{h}^{\,c}_{\mathrm{sc}}&=\mathrm{MeanPool}(\mathbf{H}^{\mathrm{sc}},m^{c}), \\
\tilde{h}^{\,c}_{\mathrm{sc}}&=\mathrm{MaxPool}(\mathbf{H}^{\mathrm{sc}},m^{c}).
\end{aligned}
\end{equation}
Here, \(\bar{h}^{\,g}_{\mathrm{sc}}\) summarizes the full valid response under the conclusion-aware scope, while \(\bar{h}^{\,c}_{\mathrm{sc}}\) and \(\tilde{h}^{\,c}_{\mathrm{sc}}\) capture the average and salient information within the conclusion span.

We further encode simple span statistics from \(m^{c}\) into a compact embedding \(\rho(m^{c})\). 
These statistics describe the relative length, log-scaled length ratio, normalized center, and normalized width of the conclusion span:
\begin{equation}\label{eq:8}
\begin{aligned}
\rho(m^{c})=\mathrm{MLP}\!\left(
\left[
\frac{|m^{c}|}{|a|};
\frac{\log(1+|m^{c}|)}{\log(1+|a|)};\right.\right. \\
\left.\left.
\frac{1}{|a|\,|m^{c}|}\sum_{t=1}^{T} t\,m_t^{c};
\frac{t_{\max}^{c}-t_{\min}^{c}+1}{|a|}
\right]
\right),
\end{aligned}
\end{equation}
where \(t_{\min}^{c}\) and \(t_{\max}^{c}\) are the first and last active positions in \(m^{c}\).

\textbf{Step 4: Reliability-aware conclusion representation.}
We next construct an initial conclusion seed \(z_0^c\) from the conclusion-centered states, the scoped global context, and the span-level cue:
\begin{equation}
\begin{aligned}
z^{c}_{0}=P_c\!\left(
\left[
\bar{h}^{\,c}_{\mathrm{sc}};
\tilde{h}^{\,c}_{\mathrm{sc}};
\bar{h}^{\,g}_{\mathrm{sc}};\right.\right. \\
\left.\left.
\rho(m^{c})+e^{c}+|\bar{h}^{\,c}_{\mathrm{sc}}-\bar{h}^{\,g}_{\mathrm{sc}}|
\right]
\right).
\end{aligned}
\end{equation}
Here, \(P_c\) is a learned projection, \(e^c\) is a conclusion-type embedding, and \(|\bar{h}^{\,c}_{\mathrm{sc}}-\bar{h}^{\,g}_{\mathrm{sc}}|\) measures the discrepancy between the conclusion and the scoped global context.

To further adapt this representation, we introduce a small learned reliability bank 
\(\mathbf{B}=[b_1,\ldots,b_4]\in\mathbb{R}^{4\times d_h}\). 
Given the initial conclusion seed \(z^{c}_{0}\), a learned scorer \(W_p\) computes the sample-specific selection weights \(\pi\) based on \(z^{c}_{0}\), the scoped global context \(\bar{h}^{\,g}_{\mathrm{sc}}\), and their discrepancy:
\begin{equation}
\begin{aligned}
\pi &= \mathrm{softmax}\!\left(W_p[z^{c}_{0};\bar{h}^{\,g}_{\mathrm{sc}};|z^{c}_{0}-\bar{h}^{\,g}_{\mathrm{sc}}|]\right), \\
p &= \pi^{\top}\mathbf{B}.
\end{aligned}
\end{equation}
Here, \(\pi\) selects reliability patterns from the bank, and the resulting sample-specific vector \(p\) is gated with the initial conclusion seed \(z^{c}_{0}\) and fused with the scoped context to obtain the reliability-aware conclusion state \(h^c\):
\begin{equation}\label{eq:11}
\begin{aligned}
g^c &= [z^{c}_{0};p;\bar{h}^{\,g}_{\mathrm{sc}};|p-z^{c}_{0}|], \\
\gamma\phantom{^{c}} &= \sigma\!\left(W_g g^c/\tau\right), \\
\tilde z^{\,c} &= R^{(1)}_c\!\left(
\gamma \odot z^{c}_{0}
+ (1-\gamma)\odot p
\right), \\
h^{c} &= R^{(2)}_c\!\left(
\mathrm{GateCtx}\!\left(\tilde z^{\,c}, \frac{p+\bar{h}^{\,g}_{\mathrm{sc}}}{2}\right)
\right).
\end{aligned}
\end{equation}
Here, \(\tau>0\) is a learned scalar temperature, \(\mathrm{GateCtx}\) is a learned context-fusion gate, and \(R^{(1)}_c\) and \(R^{(2)}_c\) are residual MLPs. 
The resulting \(h^c\) is used as the reliability-aware conclusion state for confidence prediction.

\textbf{Step 5: Intrinsic model confidence estimation.}
Finally, we fuse the global state \(h^{\,g}\) in Eq.~\ref{eq:5} with the reliability-aware conclusion state \(h^c\) in Eq.~\ref{eq:11} to estimate the intrinsic model confidence \(\boldsymbol{s}\). 
The fusion gate \(g^f\), parameterized by \(W_f\), adaptively combines global reasoning context and conclusion-specific evidence:
\begin{equation}
\begin{aligned}
g^{f}&=\sigma\!\left(W_f[h^{\,g};h^{c};|h^{\,g}-h^{c}|;h^{\,g}\odot h^{c}]\right), \\
h^{f}&=R_f\!\left(g^{f}\odot h^{c} + (1-g^{f})\odot h^{\,g}\right), \\
\boldsymbol{s}\phantom{^{f}}&=\sigma\!\left((W_oh^{f}+b_o)/T_h\right).
\end{aligned}
\end{equation}
Here, \(h^f\) is the fused representation, \(R_f\) is a residual MLP, \((W_o,b_o)\) is the output head, and \(T_h>0\) is a learned temperature that stabilizes the raw output scale. 
We train the lightweight UQ estimator \(F_{\theta}\) in eq.~\ref{eq:2} with binary cross-entropy on response correctness.
This module outputs the intrinsic model confidence \(\boldsymbol{s}\) as the raw confidence score.

\subsection{Reasoning Consistency-Aware Calibration}\label{sec:calibration}
Given the intrinsic model confidence \(\boldsymbol{s}\), this stage calibrates the estimate using reasoning-chain consistency evidence from the generated chain. 
We keep the first-stage confidence score as the calibration anchor and refine it with a fused reasoning representation that captures chain-level reliability and consistency evidence. 
For each sample, this representation is constructed as:
\begin{equation} \label{eq:13}
\begin{aligned}
u(r)=\big[
u_{\text{base}};u_{\text{reason}};u_{\text{token}};
u_{\text{gap}};u_{\text{judge}}
\big],
\end{aligned}
\end{equation}
where \(u(r)\) serves as the \textit{reasoning-chain consistency evidence} used for calibration. 
Specifically, \(u_{\text{base}}\) encodes base confidence cues, 
\(u_{\text{reason}}\) summarizes reasoning-level statistics, 
\(u_{\text{token}}\) captures token-level uncertainty signals, 
\(u_{\text{gap}}\) represents structural and feature-gap information, 
and \(u_{\text{judge}}\) encodes step-level consistency judgments.
Formal definitions are provided in Appendix~\ref{sec:app_reasoning_f}.

From a conditional reliability view, the ideal post-hoc target is the quantity introduced in Eq.~\ref{eq:3}, namely \(\mathbb{P}(y=1\mid \boldsymbol{s}, u(r))\), which refines the intrinsic model confidence \(\boldsymbol{s}\) with reasoning-derived evidence \(u(r)\) in Eq.~\ref{eq:13}. 
For brevity, we write \(u=u(r)\) and use the following anchor-plus-correction view:
\begin{equation}
\begin{aligned}
\mathbb{P}(y=1\mid \boldsymbol{s},u)
&=\underbrace{\mathbb{P}(y=1\mid \boldsymbol{s})}_{\text{anchor}}+\underbrace{\delta(\boldsymbol{s},u)}_{\text{correction}},\\
\mathbb{E}[\delta(\boldsymbol{s},u)\mid \boldsymbol{s}] &= 0.
\end{aligned}
\end{equation}
This decomposition is used as an intuitive reliability view rather than as a separate modeling assumption. The anchor represents the part of reliability already explained by the intrinsic model confidence \(\boldsymbol{s}\), while the correction term captures additional information contributed by the reasoning-chain consistency evidence \(u(r)\). When the reasoning evidence is weak or uninformative, calibration should remain close to the anchor score; when answer plausibility and chain support diverge, the correction should play a larger role. Accordingly, we construct a scalar anchor score \(\hat{\boldsymbol{s}}_{\mathrm{an}}\) from \(\boldsymbol{s}\), and a reasoning-conditioned score \(\hat{\boldsymbol{s}}_{\mathrm{re}}\) from both \(\boldsymbol{s}\) and \(u(r)\). R-Log then combines them locally across score regions. With binary cross-entropy loss 
\(\ell(y,p)=-y\log p-(1-y)\log(1-p)\), R-Log learns bin-wise mixing weights over reasoning-conditioned score regions:
\begin{equation}
\alpha_b^{*}
=\arg\min_{\alpha\in[0,1]}
\sum_{i\in\mathcal{B}_b}
\ell\!\left(y_i,\alpha\hat{\boldsymbol{s}}_{\mathrm{re},i}+(1-\alpha)\hat{\boldsymbol{s}}_{\mathrm{an},i}\right),
\end{equation}
where \(y_i\in\{0,1\}\) is the correctness label of sample \(i\), 
\(\hat{\boldsymbol{s}}_{\mathrm{an},i}\) is the scalar-calibrated anchor score, 
\(\hat{\boldsymbol{s}}_{\mathrm{re},i}\) is the reasoning-conditioned score produced by R-Log using \(u(r_i)\), 
and \(\mathcal{B}_b\) is the \(b\)-th bin induced by quantiles of \(\hat{\boldsymbol{s}}_{\mathrm{re}}\). 
Quantile binning keeps the score regions comparably populated, which makes the mixing weights less sensitive to sparse tails and improves the stability of calibration. The weight \(\alpha_b\) acts as a local trust coefficient: regions with more reliable reasoning evidence rely more on \(\hat{\boldsymbol{s}}_{\mathrm{re}}\), while other regions remain closer to the anchor score.

The calibrated confidence \(\hat{\boldsymbol{s}}_i\) is computed by adaptively combining the reasoning-conditioned score and the anchor score:
\begin{equation}
\begin{aligned}
\tilde{\boldsymbol{s}}_i &= \alpha_{b(i)}\,\hat{\boldsymbol{s}}_{\mathrm{re},i} 
+ (1-\alpha_{b(i)})\,\hat{\boldsymbol{s}}_{\mathrm{an},i},\\
\hat{\boldsymbol{s}}_i &= \mathrm{Iso}(\tilde{\boldsymbol{s}}_i).
\end{aligned}
\end{equation}
Here, \(\hat{\boldsymbol{s}}_{\mathrm{an},i}\) is the anchor score derived from the intrinsic model confidence, 
\(\hat{\boldsymbol{s}}_{\mathrm{re},i}\) is the reasoning-conditioned score estimated using the reasoning-chain consistency evidence \(u(r_i)\), 
and \(\alpha_{b(i)}\) controls their bin-wise mixture. 
The final output of this stage is the \textit{calibrated confidence score \(\hat{\boldsymbol{s}}_i\)}, which reflects both the intrinsic model confidence and the reliability of the generated reasoning chain. 
\section{Experiments}\label{sec:evaluation}
In this section, we conduct a comprehensive experimental evaluation. Specifically, we aim to address the following five research questions:

\begin{itemize}[leftmargin=5.0mm, itemsep=2pt, parsep=0pt]
    \item \textbf{RQ 1 (Performance):} How does \m compare against state-of-the-art baselines?
    \item \textbf{RQ 2 (Calibration \& Consistency):} How much do consistency signals from reasoning chains improve post-hoc calibration reliability?
    \item \textbf{RQ 3 (Cross-Distribution Generalization):} How robust is \m when applied to unseen reasoning benchmarks?
    \item \textbf{RQ 4 (Ablation Study):} What are the contributions of each component to the overall effectiveness of \m?
    \item \textbf{RQ 5 (Efficiency):} Does \m maintain a competitive computational cost?
\end{itemize}

\begin{table*}[t]
\centering
\setlength{\tabcolsep}{2pt}
\begin{adjustbox}{max width=\textwidth, keepaspectratio, max height=0.8\textheight}
\begin{tabular}{l ccc ccc ccc ccc}
\toprule
& \multicolumn{3}{c}{\textbf{HotpotQA (ID)}} & \multicolumn{3}{c}{\textbf{MuSiQue (OOD)}} & \multicolumn{3}{c}{\textbf{StrategyQA (OOD)}} & \multicolumn{3}{c}{\textbf{bAbI (OOD)}} \\
\cmidrule(lr){2-4} \cmidrule(lr){5-7} \cmidrule(lr){8-10} \cmidrule(lr){11-13}
\textbf{Method} & Acc$\uparrow$ & AUROC$\uparrow$ & PRAUC$\uparrow$ & Acc$\uparrow$ & AUROC$\uparrow$ & PRAUC$\uparrow$ & Acc$\uparrow$ & AUROC$\uparrow$ & PRAUC$\uparrow$ & Acc$\uparrow$ & AUROC$\uparrow$ & PRAUC$\uparrow$ \\
\midrule
Random & 0.494 & 0.490 & 0.738 & 0.510 & 0.509 & 0.407 & 0.512 & 0.490 & 0.857 & 0.491 & 0.492 & 0.680 \\
MCP & 0.685 & 0.533 & 0.754 & 0.427 & 0.445 & 0.368 & 0.777 & 0.429 & 0.813 & 0.683 & 0.513 & 0.711 \\
Perplexity & 0.509 & 0.469 & 0.705 & \underline{0.515} & 0.535 & 0.415 & 0.706 & 0.598 & 0.868 & 0.481 & 0.405 & 0.618 \\
MTE & 0.630 & 0.513 & 0.744 & 0.445 & 0.497 & 0.404 & 0.488 & 0.383 & 0.802 & 0.505 & 0.477 & 0.669 \\
CCP & 0.699 & 0.512 & 0.755 & 0.434 & 0.482 & 0.441 & \underline{0.852} & 0.386 & 0.836 & 0.640 & 0.494 & 0.707 \\
\midrule
SAPLMA & 0.754 & \underline{0.749} & \underline{0.861} & 0.486 & 0.604 & 0.541 & 0.835 & 0.714 & \underline{0.918} & 0.671 & 0.641 & 0.768 \\
Factoscope & 0.736 & 0.712 & 0.838 & 0.478 & \textbf{0.624} & 0.523 & 0.841 & 0.708 & 0.909 & \textbf{0.702} & \underline{0.644} & \underline{0.776} \\
Lookback & 0.750 & 0.732 & 0.852 & 0.509 & 0.615 & 0.538 & 0.839 & 0.630 & 0.876 & 0.685 & 0.640 & 0.756 \\
UHead & \underline{0.759} & 0.745 & 0.858 & 0.485 & 0.614 & \textbf{0.559} & 0.844 & \underline{0.726} & 0.916 & 0.681 & 0.625 & 0.746 \\
\rowcolor{gray!30}
\textbf{\m} & \textbf{0.779} & \textbf{0.759} & \textbf{0.884} & \textbf{0.571} & \textbf{0.625} & \underline{0.548} & \textbf{0.859} & \textbf{0.751} & \textbf{0.952} & \underline{0.694} & \textbf{0.658} & \textbf{0.788} \\
\bottomrule
\end{tabular}
\end{adjustbox}
\caption{Overall discrimination results on four datasets of Llama-3.1-8B-Instruct.}
\label{tab:result}
\end{table*}

\begin{table*}[t]
\centering
\setlength{\tabcolsep}{4pt}
\begin{adjustbox}{max width=\textwidth, keepaspectratio}
\begin{tabular}{l ccc ccc ccc ccc}
\toprule
& \multicolumn{3}{c}{\textbf{HotpotQA (ID)}} & \multicolumn{3}{c}{\textbf{MuSiQue (OOD)}} & \multicolumn{3}{c}{\textbf{StrategyQA (OOD)}} & \multicolumn{3}{c}{\textbf{bAbI (OOD)}} \\
\cmidrule(lr){2-4} \cmidrule(lr){5-7} \cmidrule(lr){8-10} \cmidrule(lr){11-13}
\textbf{Method} & ECE$\downarrow$ & Brier$\downarrow$ & Risk@90$\downarrow$ & ECE$\downarrow$ & Brier$\downarrow$ & Risk@90$\downarrow$ & ECE$\downarrow$ & Brier$\downarrow$ & Risk@90$\downarrow$ & ECE$\downarrow$ & Brier$\downarrow$ & Risk@90$\downarrow$ \\
\midrule
PS & 0.023 & 0.157 & 0.191 & 0.025 & 0.224 & 0.345 & \textbf{0.031} & 0.115 & \textbf{0.118} & \underline{0.030} & 0.197 & 0.269 \\
TS & 0.063 & 0.190 & 0.250 & 0.030 & 0.224 & 0.345 & 0.048 & 0.114 & 0.123 & 0.102 & 0.243 & 0.376 \\
ISO & \underline{0.020} & 0.157 & 0.196 & \textbf{0.017} & 0.224 & 0.346 & 0.038 & \underline{0.111} & \underline{0.120} & 0.036 & 0.197 & 0.269 \\
BWH & 0.074 & \underline{0.098} & \underline{0.088} & 0.133 & \underline{0.172} & \underline{0.242} & \textbf{0.031} & 0.115 & \textbf{0.118} & 0.112 & \textbf{0.140} & \textbf{0.141} \\
\rowcolor{gray!30}
\textbf{R-Log} & \textbf{0.011} & \textbf{0.049} & \textbf{0.041} & \underline{0.020} & \textbf{0.159} & \textbf{0.165} & \underline{0.033} & \textbf{0.109} & \textbf{0.118} & \textbf{0.020} & \underline{0.148} & \underline{0.181} \\
\bottomrule
\end{tabular}
\end{adjustbox}
\caption{Calibration results on four datasets of Llama-3.1-8B-Instruct based on \m.}
\label{tab:calibration}
\end{table*}

\subsection{Experimental Setups}

\subsubsection{Datasets}
We evaluate the performance on four widely-used public datasets: HotpotQA \citep{yang-etal-2018-hotpotqa}, MuSiQue \citep{trivedi-etal-2022-musique}, StrategyQA \citep{geva-etal-2021-strategyqa}, and bAbI \citep{weston-etal-2015-babi}. HotpotQA serves as the source in-distribution (ID) benchmark, and the other three serve as out-of-distribution (OOD) transfer benchmarks. The statistics and preprocessing details of these datasets will be shown in Appendix~\ref{sec:app_data}.

\subsubsection{Models}
Our primary experiments utilize Llama-3.1-8B-Instruct \citep{dubey2024llama3} as the frozen backbone and Mistral-Small-24B-Instruct-2501 \citep{mistral_small_24b_instruct_2025} as the judge model. Detailed specifications, extended comparisons across alternative backbones (Mistral-7B \citep{jiang2023mistral}, Gemma-2 \citep{team2024gemma2}, and Phi-4 \citep{phi4}), and a cross-judge agreement analysis using Llama-3.3-70B \citep{dubey2024llama3} and Qwen2.5-72B \citep{qwen25} are described in Appendix~\ref{sec:app_model}.

\subsubsection{Baselines}
We compare \m against both unsupervised and supervised uncertainty-aware baselines. The unsupervised group includes Random, Maximum Claim Probability (MCP)\citep{malinin2020uncertainty}, Perplexity\citep{jelinek1977perplexity}, Mean Token Entropy (MTE)\citep{malinin2020uncertainty}, and Claim Conditioned Probability (CCP)\citep{fadeeva-etal-2024-fact}. The supervised group includes SAPLMA \citep{azaria-mitchell-2023-internal-state}, Factoscope \citep{he-etal-2024-llm}, Lookback Lens \citep{chuang2024lookback}, and UHead \citep{shelmanov-etal-2025-head}. 
Detailed information will be shown in Appendix~\ref{sec:app_baseline}.

\subsubsection{Metrics}
We use three different categories of metrics: (i) response-level Accuracy; (ii) discrimination metrics such as AUROC and PRAUC; and (iii) UQ metrics including ECE, Brier score, and Risk@90, to evaluate our method. Formal definitions of all metrics are provided in Appendix~\ref{sec:app_metric}.

\subsection{Main Results (RQ 1)}
An overall comparison of \m and other baselines across four public datasets is presented in Table \ref{tab:result}. From a discrimination perspective, \m achieves the strongest Accuracy/AUROC/PRAUC values, with an average 3.1\% relative AUROC gain. Because \m and every supervised baseline are matched to roughly 2 million trainable parameters, this advantage is not explained by larger auxiliary heads. The same alignment-aware lightweight UQ stage remains competitive under OOD scenarios, preserving ranking quality without backbone retraining.
This pattern is particularly notable against stronger supervised heads, with the gains most stable on HotpotQA and StrategyQA and still competitive on the harder OOD dataset MuSiQue.

\subsection{Calibration \& Consistency (RQ 2)}
Table~\ref{tab:calibration} compares five post-hoc calibrators: Platt scaling (PS), temperature scaling (TS), isotonic regression (ISO), binwise hybrid (BWH), and Reasoning-Conditioned Logistic Calibration (R-Log). On HotpotQA, R-Log is the best on all three calibration metrics (ECE, Brier, Risk@90), and on the OOD datasets, it remains consistently competitive, especially in Brier and Risk@90. This pattern supports the design in Section~\ref{sec:methodology}: scalar calibrators mainly reshape the anchor score $\boldsymbol{s}$, whereas R-Log uses $u(r)$ to correct cases where answer plausibility and chain consistency diverge.

\subsection{Cross-Distribution Generalization (RQ 3)}

\begin{figure}[!ht]
\centering
\subfigure[Ranking vs.\ Calib gain]{\label{advantage_map}\includegraphics[width=0.49\linewidth]{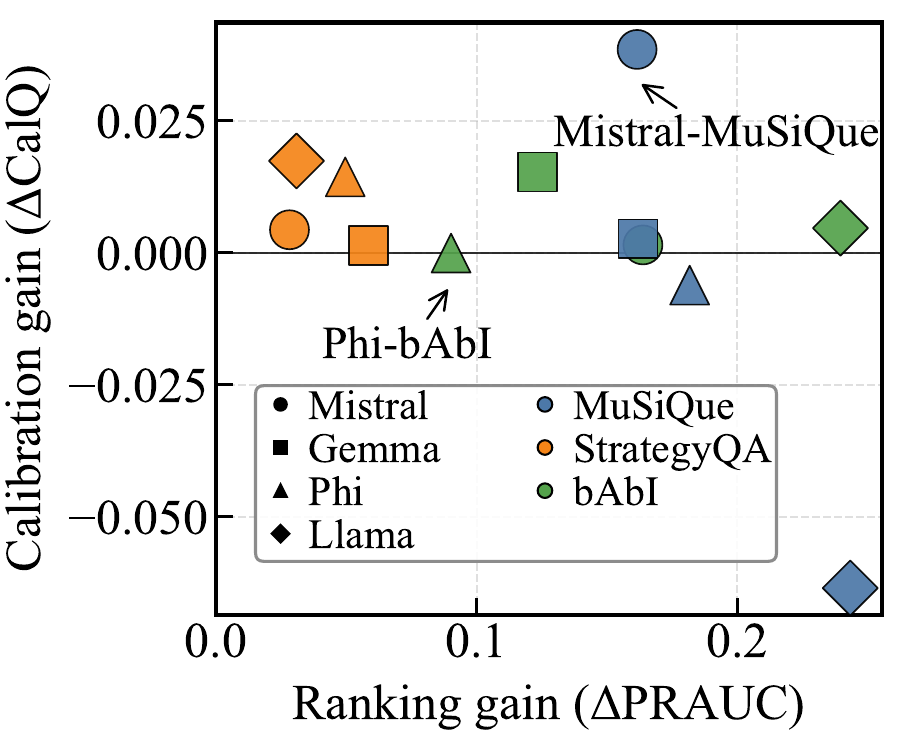}}
\hfill
\subfigure[OOD hardness vs.\ ranking]{\label{frontier}\includegraphics[width=0.49\linewidth]{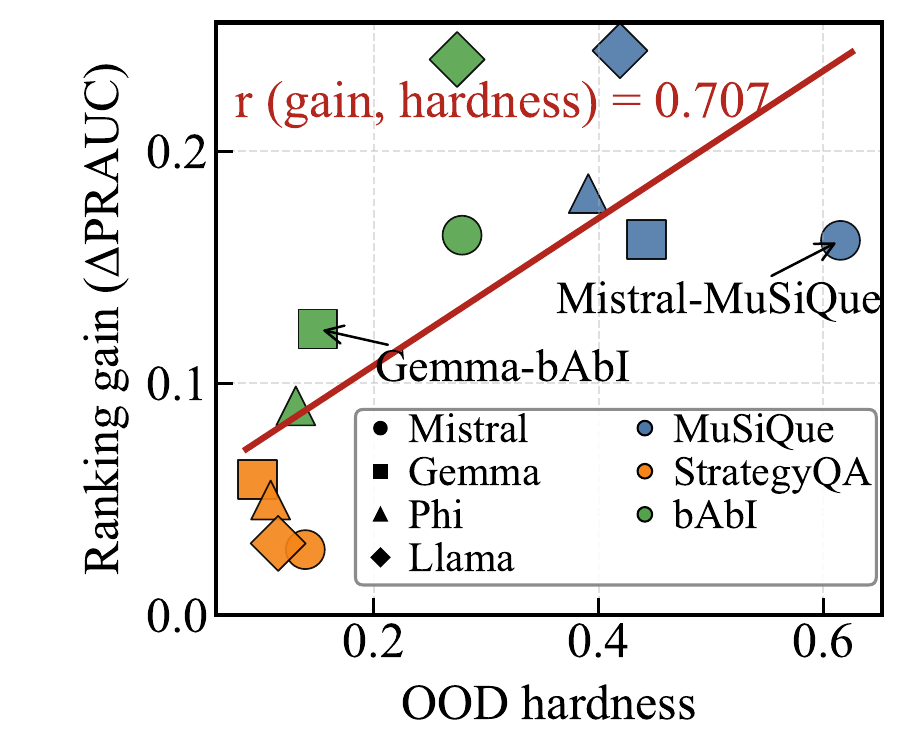}}
\caption{Cross-distribution generalization of \m. Each marker is a model-OOD pair.}
\label{fig:result}
\end{figure}

RQ3 tests whether reasoning consistency-aware calibration remains effective under distribution shifts. For each target dataset, thresholds and calibrator parameters $\phi$ (of $C_{\phi}$ in Section~\ref{sec:two_stage_uq}) are tuned on validation and transferred unchanged to test. More details of the transfer protocol are in Appendix~\ref{sec:app_transfer_protocol}.
As shown in Figure~\ref{fig:result}, the horizontal axis is ranking gain:
\begin{equation*}
\Delta \text{PRAUC}=\text{PRAUC}_{\text{R-Log}}-\text{PRAUC}_{\text{ISO}},
\end{equation*}
and the vertical axis is calibration-quality gain:
\begin{align*}
\Delta \text{CalQ}
&=(1-\text{ECE}_{\text{R-Log}})-(1-\text{ECE}_{\text{ISO}}) \\
&=\text{ECE}_{\text{ISO}}-\text{ECE}_{\text{R-Log}}.
\end{align*}
Positive $\Delta \text{PRAUC}$ or $\Delta \text{CalQ}$ means R-Log outperforms ISO on the corresponding axis. In the right panel, OOD hardness is defined as $1-\text{PRAUC}_{\text{ISO}}$, where $\text{PRAUC}_{\text{ISO}}$ is ISO's PRAUC on that OOD benchmark.

The key observation from Figure~\ref{advantage_map} is that gains are mostly joint: most model--OOD pairs move toward better discrimination and better calibration at the same time, rather than exchanging one for the other. 
Figure~\ref{frontier} adds a difficulty perspective: as OOD hardness ($1-\text{PRAUC}_{\text{ISO}}$) increases, the ranking gain of R-Log over ISO tends to grow (Pearson $r=0.707$). Table~\ref{tab:calibration} is consistent with the same transfer story numerically: on harder OOD settings, R-Log continues to improve Brier score and remains competitive or better on Risk@90 relative to ISO.

\subsection{Ablation Study (RQ 4)}
\begin{table}[t]
\centering
\setlength{\tabcolsep}{2pt}
\begin{adjustbox}{max width=\linewidth, keepaspectratio}
\begin{tabular}{l ccc ccc}
\toprule
& \multicolumn{3}{c}{\textbf{HotpotQA (ID)}} & \multicolumn{3}{c}{\textbf{MuSiQue (OOD)}} \\
\cmidrule(lr){2-4} \cmidrule(lr){5-7} 
\textbf{Variant} & ECE$\downarrow$ & Brier$\downarrow$ & Risk@90$\downarrow$ & ECE$\downarrow$ & Brier$\downarrow$ & Risk@90$\downarrow$ \\
\midrule
\rowcolor{gray!30}
\textbf{Full} & \textbf{0.011} & \textbf{0.049} & \textbf{0.041} & \textbf{0.020} & \textbf{0.159} & \textbf{0.165} \\
-LB & 0.014 & 0.052 & 0.045 & 0.107 & 0.185 & 0.245 \\
-CS & 0.016 & 0.055 & 0.048 & 0.098 & 0.181 & 0.252 \\
-CF & \underline{0.013} & 0.053 & \underline{0.044} & \underline{0.088} & 0.193 & 0.270 \\
-FR & 0.018 & \underline{0.051} & 0.047 & 0.094 & \underline{0.177} & \underline{0.238} \\
\bottomrule
\end{tabular}
\end{adjustbox}
\caption{Ablation of the UQ-head Modules.}
\label{tab:ablation}
\end{table}
Table~\ref{tab:ablation} reports four ablations of the modules introduced in Section~\ref{sec:methodology}. We use \textbf{-LB}, \textbf{-CS}, \textbf{-CF}, and \textbf{-FR} for removing the learned bank, conditioned selection, context fusion, and final residual refiner, respectively.
On the ID dataset HotpotQA and the more difficult OOD dataset MuSiQue, every simplification degrades calibration, suggesting that these refinement steps are complementary rather than redundant. \textbf{-LB} causes the largest ECE deterioration, while \textbf{-CF} causes the largest Brier and Risk@90 degradation. \textbf{-CS} and \textbf{-FR} are also consistently worse than Full, indicating that each component contributes to robust calibration.

\subsection{Efficiency (RQ 5)}
Table~\ref{tab:efficiency} and Table~\ref{tab:efficiency_post} in Appendix~\ref{sec:app_hardware_eff} report the inference efficiency for both the Stage 1 UQ heads and the Stage 2 post-hoc calibration. 
For the UQ heads, unsupervised token-level scores are naturally much faster than learned heads. Among supervised baselines, Factoscope and Lookback are the fastest, whereas UHead is the most memory-intensive. Although our best-performing \m head on HotpotQA is marginally slower, it operates in the same post-generation regime and successfully reduces peak memory relative to UHead. 
Furthermore, for post-hoc calibration, R-Log demonstrates a better balance between effectiveness and speed, remaining much lighter than BWH while still preserving reasoning-aware calibration.
\section{Related Work}\label{sec:related-work}

\subsection{Uncertainty Quantification Heads}
Existing uncertainty quantification (UQ) methods for LLMs can be broadly categorized into unsupervised and supervised approaches. Unsupervised methods estimate uncertainty via predictive statistics, including confidence scores like MCP \citep{malinin2020uncertainty} and likelihood surrogates like Perplexity \citep{jelinek1977perplexity} and CCP \citep{fadeeva-etal-2024-fact}. Supervised approaches instead train auxiliary predictors over internal states, encompassing internal-state probes such as SAPLMA \citep{azaria-mitchell-2023-internal-state} and UHead \citep{shelmanov-etal-2025-head}, or reasoning-oriented methods like Factoscope \citep{he-etal-2024-llm} and Lookback Lens \citep{chuang2024lookback}.

Despite their effectiveness, most existing supervised heads estimate uncertainty from aggregated representations without explicitly separating the final conclusion from the supporting reasoning process. In contrast, \m decomposes uncertainty estimation into conclusion-focused prediction and reasoning consistency-aware calibration. This design separates lightweight conclusion-aligned scoring from subsequent reasoning-based reliability adjustment.

\subsection{Reasoning-Aware Uncertainty Estimation}

Beyond probing-based methods, uncertainty in LLMs has also been studied through verbalized confidence \citep{kadavath-etal-2022-know-what,yona-etal-2024-intrinsic-uncertainty}, black-box disagreement signals \citep{manakul-etal-2023-selfcheckgpt}, and semantic invariance measures \citep{kuhn-etal-2023-semantic-uncertainty,kossen-etal-2024-semantic-entropy-probes}. Reasoning-oriented benchmarks, including StrategyQA and multi-hop or spatial reasoning tasks \citep{geva-etal-2021-strategyqa,ferguson-etal-2020-iirc,ho-etal-2020-2wiki,mirzaee-etal-2021-spartqa,shi-etal-2022-stepgame}, further highlight the importance of reliable confidence estimation under heterogeneous reasoning distributions. Meanwhile, process-supervision and verifier-based studies demonstrate the value of intermediate reasoning signals for reliability assessment \citep{wei-etal-2022-chain-of-thought,cobbe-etal-2021-training-verifiers,lightman-etal-2023-verify-step,min-etal-2023-factscore}.

Unlike verifier-based reranking or answer selection approaches that use reasoning signals to modify outputs, \m treats reasoning consistency as post-hoc calibration evidence while preserving the original prediction. Different from conventional one-dimensional calibration methods, \m augments uncertainty calibration with reasoning consistency signals while maintaining efficient single-trace inference. Overall, \m combines internal-state probing with reasoning consistency-aware post-hoc calibration for uncertainty estimation in LLM reasoning tasks.

\section{Conclusion}\label{sec:conclusion}

In this work, we propose \m, a reasoning consistency-aware UQ framework for LLMs. \m includes two novel designs, i.e., an alignment-aware lightweight UQ module that estimates a raw intrinsic model confidence score from frozen features aligned to the final conclusion, and a reasoning consistency-aware calibration module that calibrates this score with reasoning-chain consistency evidence. Extensive experiments on four datasets show that \m achieves an average 3.1\% relative gain in AUROC and robust ID--OOD transfer without additional fine-tuning.
Calibration results further show that reasoning evidence is valuable when scalar correction is insufficient, achieving up to 45.0\% relative reduction in ECE. Overall, \m improves response-level UQ without larger auxiliary heads or repeated generation.

\clearpage
\section*{Limitations}\label{sec:limitations}
\m has three practical limitations. First, it is a single-trace, post-generation pipeline over judged caches, so the strongest reasoning consistency-aware calibrators still depend on a verification stage and inherit noise or bias from the judge model; in our design, this primarily affects calibration because step-level consistency labels never directly supervise the response predictor. Second, \m does not repair generation errors or search over alternative chains: when the base model produces weakly structured or question-misaligned reasoning, the conclusion-aligned features become less informative and transfer is harder. Third, the main paper emphasizes a fully comparable English-language evaluation on which all systems are completed under one protocol, so broader multilingual, interactive, or retrieval-augmented settings remain future work.

\section*{Ethical Considerations}\label{sec:ethics}
All authors have read and agree to abide by the ACL Code of Ethics. This work uses publicly available datasets and does not involve private or personally identifying information, but confidence scores and judged reasoning labels can still be misused if treated as guarantees of correctness, truthfulness, or safety. In our view, \m should be used as a decision-support signal for abstention, triage, and human review rather than as a substitute for expert verification, especially in high-stakes settings. Because the pipeline also depends on an external judge model, any downstream deployment should account for possible annotation bias, overconfidence, and domain mismatch in both the generator and the verifier.

\bibliography{reference}

\clearpage
\appendix

\section*{Appendix}
\label{sec:appendix}

\section{Pipeline Details}\label{sec:app_pipeline}

\m is a \emph{post-generation}, single-trace reasoning UQ pipeline over frozen internal states. The backbone LLM is kept fixed, and the operational target is a \textbf{response-level} confidence score attached to the final conclusion after decoding. Throughout the appendix, we follow the same notation as in the main paper: the alignment-aware lightweight UQ module produces an intrinsic model confidence score \(\boldsymbol{s}\in[0,1]\), the reasoning consistency-aware calibrator refines it into a calibrated confidence score \(\hat{\boldsymbol{s}}\in[0,1]\), and the per-sample calibrated score is written as \(\hat{\boldsymbol{s}}_i\). The reasoning-chain calibration evidence is summarized as \(u(r)\).

\textbf{1. Input \& Generation.} Given a question and its corresponding context, the frozen backbone white-box LLM generates a Chain-of-Thought (CoT) reasoning chain. This chain is segmented into atomic reasoning steps (e.g., Step 1, Step 2) and culminates in a final conclusion claim within the generated response.

\textbf{2. Judge.} A separate verification protocol evaluates the generated response. This process consists of two parts: reasoning consistency verification to evaluate the validity and logical flow of the individual CoT steps, and answer verification to check the final conclusion. This judge yields a ground-truth binary training signal $y \in \{0,1\}$ (where 0 represents incorrect and 1 represents correct) alongside structural consistency labels for the generated reasoning steps, as introduced in Section~\ref{sec:consistency-aware}.

\textbf{3. Feature Extraction.} Aligned with the generated tokens, internal states are extracted per-token and per-layer from the frozen LLM. The derived features written to the disk cache primarily include hidden representations, attention maps, and token probabilities. These compact tensors are indexed and remain unchanged for downstream consumption.

\textbf{4. Training.} The training process is divided into two sequential stages. First, the Alignment-Aware Lightweight UQ Head is trained on the extracted features with the binary correctness signal \(y\) to produce the intrinsic model confidence score \(\boldsymbol{s}\), which serves as the first-stage raw confidence. Second, the Reasoning Consistency-Aware Calibrator (\(C_{\phi}\)) is trained to refine this raw score. By incorporating a rich reasoning feature vector \(u(r)\), which encapsulates base-score features, reasoning summaries, token-level uncertainty, structure/feature-gap statistics, and judged reasoning consistency, the reliability estimate is refined to yield the calibrated confidence \(\hat{\boldsymbol{s}} = C_{\phi}(\boldsymbol{s}, u(r))\).

\textbf{5. Inference.} During test-time deployment, the pipeline operates end-to-end. A new question prompts the generation of a CoT trace and conclusion. Internal features are extracted and routed through the trained UQ Head to yield \(\boldsymbol{s}\). For the full calibrated setting used in our main experiments, the same verification stage also provides the judged reasoning metadata needed to instantiate \(u(r)\), after which the Calibrator applies the learned refinements to output the final calibrated confidence \(\hat{\boldsymbol{s}} \in [0,1]\). The system ultimately outputs both the generated conclusion and its calibrated confidence score.

Judge-based verification is used only where a lexical exact match is not an adequate measurement layer. StrategyQA remains exact-match because its binary answer space can be decomposed deterministically. By contrast, HotpotQA, MuSiQue, and our bAbI setup permit short free-form responses whose wording may vary despite equivalent semantics. We therefore apply one stronger judge uniformly as a shared measurement layer for response correctness and step-level consistency on those tasks. The judge sees only task inputs and generated text, not the originating baseline or confidence score, so this verification layer does not create a method-specific advantage for \m.

\begin{table*}[t]
\centering
\setlength{\tabcolsep}{4pt}
\begin{adjustbox}{max width=\textwidth, keepaspectratio}
\begin{tabular}{l cc cc cc cc}
\toprule
& \multicolumn{2}{c}{\textbf{HotpotQA}} & \multicolumn{2}{c}{\textbf{MuSiQue}} & \multicolumn{2}{c}{\textbf{StrategyQA}} & \multicolumn{2}{c}{\textbf{bAbI}} \\
\cmidrule(lr){2-3} \cmidrule(lr){4-5} \cmidrule(lr){6-7} \cmidrule(lr){8-9}
\textbf{Verification Target} & Agree. (\%)$\uparrow$ & $\kappa\uparrow$ & Agree. (\%)$\uparrow$ & $\kappa\uparrow$ & Agree. (\%)$\uparrow$ & $\kappa\uparrow$ & Agree. (\%)$\uparrow$ & $\kappa\uparrow$ \\
\midrule
\multicolumn{9}{l}{\textit{Reference Judge: Qwen2.5-72B-Instruct}} \\
\midrule
Final conclusion correctness & 95.8 & 0.88 & 94.4 & 0.85 & 98.1 & 0.94 & 99.2 & 0.97 \\
Step-level reasoning consistency & 93.5 & 0.84 & 91.7 & 0.80 & 96.3 & 0.90 & 98.0 & 0.95 \\
\midrule
\multicolumn{9}{l}{\textit{Reference Judge: Llama-3.3-70B-Instruct}} \\
\midrule
Final conclusion correctness & 96.2 & 0.89 & 94.9 & 0.86 & 98.5 & 0.95 & 99.5 & 0.98 \\
Step-level reasoning consistency & 94.0 & 0.85 & 92.2 & 0.81 & 96.8 & 0.91 & 98.3 & 0.96 \\
\bottomrule
\end{tabular}
\end{adjustbox}
\caption{Cross-judge agreement analysis comparing Mistral-Small-24B against stronger reference judges. ``Agree.'' denotes agreement rate, and $\kappa$ denotes Cohen's kappa.}
\label{tab:judge_agreement}
\end{table*}

\section{Experimental Setups}\label{sec:app_experiment}
\subsection{Datasets and Preprocessing}\label{sec:app_data}
We evaluate the performance using four benchmarks that emphasize different aspects of reasoning. \textbf{HotpotQA} \citep{yang-etal-2018-hotpotqa} serves as the source benchmark. \textbf{MuSiQue} \citep{trivedi-etal-2022-musique} stresses cross-document evidence aggregation, \textbf{StrategyQA} \citep{geva-etal-2021-strategyqa} emphasizes implicit reasoning and commonsense composition, and \textbf{bAbI} \citep{weston-etal-2015-babi} provides a more controlled symbolic reasoning environment.

We use the official train, validation, and test splits whenever they are provided by the original dataset authors. HotpotQA provides the source training split; the remaining three datasets are used for transfer validation and test. Full split statistics are listed in Table~\ref{tab:dataset}.
\begin{table}[ht]
\centering
\footnotesize
\begin{adjustbox}{max width=\linewidth, keepaspectratio}
\begin{tabular}{l cccc}
\toprule
\textbf{Dataset}& \textbf{Type} & \textbf{Train} & \textbf{Val} & \textbf{Test} \\ 
\midrule
 HotpotQA      & Generation   & 90,447 & 3,702 & 3,703 \\
 MuSiQue       & Generation   & 19,938 & 1,208 & 1,209 \\
 StrategyQA    & Choice       & 1,443  & 160   & 687 \\
 bAbI          & Generation   & 18,013 & 1,987 & 20,000 \\
\bottomrule
\end{tabular}
\end{adjustbox}
\caption{Full dataset split sizes used in this work.}
\label{tab:dataset}
\end{table}

To evaluate predictions, we use task-dependent scoring. For exact-match tasks such as StrategyQA, the final conclusion is extracted with deterministic rules and matched directly against the label. For free-form tasks such as HotpotQA, MuSiQue, and bAbI, we rely on Mistral-Small-24B-Instruct-2501~\citep{mistral_small_24b_instruct_2025} as an external judge for response correctness and reasoning verification.

All four datasets are publicly released research artifacts, and we use them under their original public licenses or terms of use. This is consistent with the intended use: we evaluate the original reasoning tasks on the official splits and do not redistribute modified dataset copies. We also inspected the fields and example contents consumed by our pipeline, including question, context/passage, facts, decomposition, and answers, and did not find personally identifying information or systematic offensive content.

\subsection{Models and Agreement Analysis}\label{sec:app_model}

We evaluate the pipeline on four target backbones: \textbf{Llama-3.1-8B-Instruct} \citep{dubey2024llama3}, \textbf{Mistral-7B-Instruct-v0.3} \citep{jiang2023mistral}, \textbf{Gemma-2-9B-it} \citep{team2024gemma2}, and \textbf{Phi-4}. All target models remain frozen during the uncertainty estimation process. \textbf{Mistral-Small-24B-Instruct-2501} \citep{mistral_small_24b_instruct_2025} is primarily used as the judge model for open-ended conclusion grading and intra-chain reasoning consistency verification. \textbf{Llama-3.3-70B-Instruct} \citep{dubey2024llama3} and \textbf{Qwen2.5-72B-Instruct} \citep{qwen25} serve as stronger reference judges for our cross-judge agreement analysis. Table~\ref{tab:models} summarizes the model sizes and their respective roles.

\begin{table}[ht]
\centering
\small
\begin{tabular}{l cc}
\toprule
\textbf{Model} & \textbf{Params} & \textbf{Role}\\ 
\midrule
 Llama-3.1-8B-Instruct      & 8B  & Target\\
 Mistral-7B-Instruct-v0.3   & 7B  & Target\\
 Gemma-2-9B-it              & 9B  & Target\\
 Phi-4                      & 14B & Target\\
 Mistral-Small-24B-Instruct-2501 & 24B & Judge\\ 
 Llama-3.3-70B-Instruct     & 70B & Judge\\ 
 Qwen2.5-72B-Instruct       & 72B & Judge\\
\bottomrule
\end{tabular}
\caption{Models used in our experiments. Target models generate reasoning traces and provide frozen features for uncertainty estimation; the judge models are used for answer grading and reasoning verification.}
\label{tab:models}
\end{table}

In our main pipeline, we utilize Mistral-Small-24B-Instruct-2501 as the judge model to balance evaluation quality with computational efficiency. To rigorously validate that this lightweight choice does not introduce significant annotation bias or compromise the integrity of our reasoning consistency signals, we conduct an ablation study using the much larger Qwen2.5-72B-Instruct and Llama-3.3-70B-Instruct as stronger reference judges. 
We re-evaluate the same target-model responses with the 70B+ judges using the identical verification prompts. As shown in Table~\ref{tab:judge_agreement}, the 24B judge achieves a highly robust agreement rate ($>91\%$) and strong Cohen's $\kappa$ ($\ge 0.80$) with both reference judges across all datasets for both final conclusion correctness (Stage 1) and step-level reasoning consistency (Stage 2). 
This high alignment confirms that the 24B judge provides reliable consistency signals for our framework, even on complex multi-hop reasoning tasks like MuSiQue. Consequently, using the 24B model is a sound methodological choice that significantly reduces the preprocessing footprint without sacrificing the quality of the reasoning consistency-aware calibration.

\subsection{Baselines}\label{sec:app_baseline}

To rigorously evaluate our proposed framework, we first compare it against \textbf{unsupervised} baselines that compute uncertainty directly from the model's output probability distributions without additional training. As a naive lower bound for AUROC and PRAUC metrics, we include a \textbf{Random} baseline that assigns uniform scores to generated responses. We also evaluate standard confidence estimation metrics, including \textbf{Maximum Claim Probability (MCP)} \citep{malinin2020uncertainty, fadeeva-etal-2024-fact}, which we adapt to response-level scoring by aggregating the strongest extracted claim confidence, and \textbf{Perplexity (PPL)} \citep{jelinek1977perplexity}, which reflects the model's overall generative confidence based on mean negative log-likelihood. To capture token-level predictive uncertainty across the sequence, we utilize \textbf{Mean Token Entropy (MTE)} \citep{malinin2020uncertainty}, which measures the average Shannon entropy at each autoregressive decoding step. Finally, we apply \textbf{Claim Conditioned Probability (CCP)} \citep{fadeeva-etal-2024-fact}, a token-level metric that isolates factual uncertainty from linguistic variation by measuring the probability of a candidate value conditioned on the context and then applying a CCP-style reduction. 

In contrast, \textbf{supervised} baselines train lightweight classifiers on frozen internal activations of the target LLMs. \textbf{SAPLMA} \citep{azaria-mitchell-2023-internal-state} serves as a simple feed-forward probe over hidden representations. \textbf{Factoscope} \citep{he-etal-2024-llm} emphasizes discrepancies between local and global internal scopes. \textbf{Lookback Lens} \citep{chuang2024lookback} emphasizes contextual signals derived from internal attention patterns. Finally, \textbf{UHead} \citep{shelmanov-etal-2025-head} represents plug-and-play auxiliary uncertainty heads. In our pipeline, all supervised baselines are instantiated as response-level confidence estimators over the same frozen internal states and evaluated under the same calibration protocol as the main method.

The baseline artifacts used for comparison are likewise public research artifacts. When an official public implementation is available, we use it under its original open license or usage terms; otherwise, we implement the published method inside our framework from the paper description. This is also consistent with the intended use, because the baselines are only used for transparent research comparison on the same public benchmarks.

\subsection{Metrics}\label{sec:app_metric}

We define here only the metrics reported in the main paper: Accuracy, AUROC, PRAUC, ECE, Brier, and Risk@90. Let \(N\) be the total number of evaluated samples. For the \(i\)-th sample, let \(y_i \in \{0, 1\}\) denote the ground-truth response-level correctness label, let \(\hat{\boldsymbol{s}}_i \in [0, 1]\) denote the final calibrated confidence score produced by the uncertainty quantification method, let \(\hat{y}_i\in\{0,1\}\) denote the predicted correctness label after the evaluation thresholding protocol, let \(\hat{y}_i^{(0.5)}=\mathbf{1}[\hat{\boldsymbol{s}}_i\ge \frac{1}{2}]\) denote the score-induced correctness prediction, and let \(c_i=\max(\hat{\boldsymbol{s}}_i,1-\hat{\boldsymbol{s}}_i)\) denote the associated classification confidence. When the appendix refers to the pre-calibration score, we use the same raw-confidence notation \(\boldsymbol{s}\) as in Eq.~\ref{eq:2}. All metric outputs are unitless scalars, and lower values are better for ECE/Brier/Risk@90 while higher values are better for Accuracy/AUROC/PRAUC.

\subsubsection{Thresholded Classification Metric}
\begin{itemize}[leftmargin=*, itemsep=2pt, parsep=0pt]
    \item \textbf{Accuracy (Acc):} Measures the agreement between the predicted correctness labels and the ground-truth correctness labels.
    \begin{equation}
    \text{Acc} = \frac{1}{N} \sum_{i=1}^N \mathbf{1}[\hat{y}_i = y_i]
    \end{equation}
\end{itemize}

\subsubsection{Discrimination Metrics}
Discrimination metrics evaluate how well the estimated confidence scores \(\hat{\boldsymbol{s}}_i\) can separate correct predictions from incorrect ones, independently of the absolute scale of \(\hat{\boldsymbol{s}}_i\).
\begin{itemize}[leftmargin=*, itemsep=2pt, parsep=0pt]
    \item \textbf{AUROC (Area Under the Receiver Operating Characteristic curve):} Represents the probability that a randomly selected correct response is assigned a higher confidence score than a randomly selected incorrect response.
    \begin{equation}
    \text{AUROC} = \Pr(\hat{\boldsymbol{s}}_i > \hat{\boldsymbol{s}}_j \mid y_i = 1, y_j = 0)
    \end{equation}
    where $i$ and $j$ are sample indices drawn from the positive and negative sets respectively, and $\Pr(\cdot)$ denotes probability under this pairwise draw.
    \item \textbf{PRAUC (Area Under the Precision-Recall curve):} Summarizes the precision-recall trade-off across all possible confidence thresholds. Let $\text{Prec}(r)$ denote the precision at a given recall level $r \in [0, 1]$ as the decision threshold varies. The metric is computed as the integral of precision with respect to recall:
    \begin{equation}
    \text{PRAUC} = \int_{0}^{1} \text{Prec}(r) \, dr
    \end{equation}
    This metric is particularly informative when the distribution of correct and incorrect responses is highly skewed.
\end{itemize}

\subsubsection{Calibration and Selective-Risk Metrics}
Calibration metrics quantify the alignment between the score-induced classification confidence $c_i$ and empirical correctness, while selective-risk metrics measure how reliably the model can retain only its most confident predictions. For ECE, we partition the $N$ samples into $M$ bins. Let $B_m$ be the set of indices of samples in the $m$-th bin. The empirical accuracy and average confidence for bin $B_m$ are defined as $\text{acc}(B_m) = \frac{1}{|B_m|} \sum_{i \in B_m} \mathbf{1}[\hat{y}_i^{(0.5)} = y_i]$ and $\text{conf}(B_m) = \frac{1}{|B_m|} \sum_{i \in B_m} c_i$, respectively, where $\lvert B_m\rvert$ is the number of samples in bin $m$.
\begin{itemize}[leftmargin=*, itemsep=2pt, parsep=0pt]
    \item \textbf{ECE (Expected Calibration Error):} Partitions the confidence interval $[0, 1]$ into $M$ equally spaced bins and calculates the expected difference between empirical correctness and confidence, weighted by the number of samples in each bin.
    \begin{equation}
    \text{ECE} = \sum_{m=1}^M \frac{|B_m|}{N} \big| \text{acc}(B_m) - \text{conf}(B_m) \big|
    \end{equation}
    \item \textbf{Brier Score:} Measures the mean squared difference between the predicted confidence and the binary correctness target, so it captures both calibration quality and sharpness.
    \begin{equation}
    \text{Brier} = \frac{1}{N} \sum_{i=1}^{N} (\hat{\boldsymbol{s}}_i - y_i)^2
    \end{equation}
    \item \textbf{Risk@90:} Sorts samples by classification confidence, keeps the top 90\% most confident predictions, and reports the residual error rate on that subset. Let $\pi$ be a permutation of $\{1,\ldots,N\}$ such that $c_{\pi_1} \ge \cdots \ge c_{\pi_N}$, and let $S_{0.9}=\{\pi_1,\ldots,\pi_{\lceil 0.9N \rceil}\}$.
    \begin{equation}
    \text{Risk@90} = 1 - \frac{1}{|S_{0.9}|} \sum_{i \in S_{0.9}} \mathbf{1}[\hat{y}_i^{(0.5)} = y_i]
    \end{equation}
\end{itemize}

\subsection{Reasoning-Aware Calibration Features}\label{sec:app_reasoning_f}

For a generated trace \(r\), let \(z\) be the sample-level pre-sigmoid logit corresponding to the raw confidence \(\boldsymbol{s}=\sigma(z)\), let \(z^c\) and \(p^c=\sigma(z^c)\) be the conclusion-claim logit and probability, and let \((\ell_k,p_k)\), \(k=1,\dots,K\), denote the usable reasoning-claim logits and probabilities. We further use \(\lambda_t=\log p(y_t\mid y_{<t},x)\) for the realized token log-likelihood at decoding step \(t\), \(q_t^{(1)}\ge q_t^{(2)}\) for the top-1 and top-2 token log-probabilities, \(h\) for the scalar task-difficulty indicator (e.g., hop count or a dataset-specific difficulty field), and \(y_k^{r}\in\{0,1\}\), \(k=1,\dots,K_v\), for the valid judged reasoning labels returned by the stage-2 reasoning-consistency verifier.

The reasoning-aware calibration feature vector is
\begin{equation}
\begin{aligned}
u(r)=\big[
u_{\text{base}};u_{\text{reason}};u_{\text{token}};
u_{\text{gap}};u_{\text{judge}}
\big].
\end{aligned}
\end{equation}
For brevity, define
\begin{equation}
\begin{aligned}
\operatorname{Stat}(a_{1:n})=\big[&
\operatorname{Mean}(a);\min(a);\\
&\max(a);\operatorname{Std}(a)
\big].
\end{aligned}
\end{equation}
We keep three count symbols distinct: $K$ is the number of usable reasoning-score entries, $K_c$ is the segmented reasoning-claim count, and $K_v$ is the number of valid judged reasoning labels. They are often equal on clean samples, but may differ after filtering or failed judge parses. In preprocessing, we retain only samples with at least one usable reasoning score, one segmented reasoning claim, and one valid judged reasoning label, so the denominators in the following formulas satisfy $K, K_c, K_v \ge 1$.

\paragraph{1) Base-score features.}
\begin{equation}
u_{\text{base}}=[z;z^c;p^c].
\end{equation}

\paragraph{2) Reasoning-summary features.}
\begin{equation}
\begin{aligned}
\bar{p}&=\frac{1}{K}\sum_{k=1}^{K}p_k,\\
u_{\text{reason}}=\big[&
\operatorname{Stat}(\ell_{1:K});
\operatorname{Stat}(p_{1:K});\\
&\operatorname{Var}(p_{1:K});
\lvert p^c-\bar{p}\rvert;\\
&z-z^c;K
\big].
\end{aligned}
\end{equation}
We retain $\operatorname{Var}(p_{1:K})$ in addition to $\operatorname{Std}(p_{1:K})$ as a separate scale-sensitive dispersion feature for the downstream logistic calibrator.

\paragraph{3) Token-uncertainty features.}
\begin{equation}
\begin{aligned}
u_{\text{token}}=\big[&
-\frac{1}{T}\sum_{t=1}^{T}\lambda_t; -\frac{1}{T}\sum_{t=1}^{T} q_t^{(1)};\\
&-\frac{1}{T}\sum_{t=1}^{T}\big(q_t^{(1)}-q_t^{(2)}\big)
\big].
\end{aligned}
\end{equation}

\paragraph{4) Structure and feature-gap features.}
Let $\bar{f}^{\,c}$ and $\bar{f}^{\,r}$ denote the mean-pooled conclusion and reasoning features, let $g_j\in\mathbb{R}^{d_f}$, $j=1,\dots,T_r$, denote the token features extracted from the cached reasoning span (the reasoning sidecar), let $\delta_f=\bar{f}^{\,r}-\bar{f}^{\,c}$, let $T_r$ be the number of reasoning-sidecar tokens, let $T_r^+$ be the number of active reasoning tokens, and let $K_c$ be the segmented reasoning-claim count. Define
\begin{equation}
\begin{aligned}
\operatorname{Gap}(\delta_f)=\big[&
\|\delta_f\|_2;\frac{1}{d_f}\|\delta_f\|_1;\\
&\|\delta_f\|_{\infty};\cos(\bar{f}^{\,r},\bar{f}^{\,c})
\big].
\end{aligned}
\end{equation}
and
\begin{equation}
\begin{aligned}
\operatorname{Mag}(g)=\big[&
T_r;T_r^+;\\
&\operatorname{Mean}_j\|g_j\|_2;
\operatorname{Std}_j\|g_j\|_2;\\
&\operatorname{Mean}_{j,m}|g_{j,m}|;
\operatorname{Std}_{j,m}|g_{j,m}|
\big].
\end{aligned}
\end{equation}
Then
\begin{equation}
\begin{aligned}
u_{\text{gap}}=\big[&
h;K_c;\frac{T_r}{K_c};\frac{T_r^+}{K_c};\\
&\operatorname{Gap}(\delta_f);\operatorname{Mag}(g)
\big].
\end{aligned}
\end{equation}

\paragraph{5) Judged reasoning features.}
\begin{equation}
\begin{aligned}
u_{\text{judge}}=\big[&
\frac{1}{K_v}\sum_{k=1}^{K_v} y_k^{r};
\operatorname{Std}(y_{1:K_v}^{r});\\
&\frac{1}{K_v}\sum_{k=1}^{K_v}\mathbf{1}\!\left[y_k^{r}=
\mathbf{1}\!\left[\frac{1}{K_v}\sum_{j=1}^{K_v}y_j^{r}\ge\frac{1}{2}\right]\right];\\
&\left|\frac{1}{K_v}\sum_{k=1}^{K_v} y_k^{r}-\frac{1}{2}\right|;
K_v
\big].
\end{aligned}
\end{equation}
These judge features depend only on the stage-2 reasoning-verification outputs and their internal agreement statistics; they do not use the response-level correctness label $y$ (i.e., the stage-1 answer-correctness verdict) as an input feature. This avoids label leakage and keeps the feature definition consistent with the post-generation inference setting in Section~\ref{sec:app_pipeline}.

Overall, $u(r)$ combines base confidence cues, reasoning-claim summaries, token-level uncertainty, structure/feature-gap statistics, and judge-derived consistency signals.

\subsection{Transfer Protocol}\label{sec:app_transfer_protocol}
All decision thresholds and calibration parameters are fit on a held-out validation split and then frozen before the corresponding test evaluation. For the source benchmark, the HotpotQA validation split is used to tune the thresholding protocol and, when enabled, the calibration map. The resulting configuration is then transferred unchanged to the HotpotQA test split. For out-of-distribution evaluation, the same procedure is repeated on each target dataset: its validation split is used for score fitting, and the learned threshold/calibrator is applied once to the corresponding test split with no further adjustment.

This separation is import ant because our goal is to evaluate transferable response-level uncertainty estimation, not test-set-specific retuning. The first-stage predictor is fixed after training, and only the held-out validation data is allowed to adapt the scalar threshold or the optional reasoning consistency-aware calibration layer. As a result, the reported OOD numbers reflect whether the same two-stage unified representation remains informative when reasoning distribution, evidence structure, and answer format shift across datasets, rather than hidden adaptation on the final test distribution.

Unless otherwise stated, every quantitative result reported in the main paper and this appendix is the arithmetic mean over 10 runs with different random seeds under the same train/validation/test protocol; we do not report cherry-picked best single runs.

\begin{table}[t]
\centering
\setlength{\tabcolsep}{2pt}
\begin{adjustbox}{max width=\linewidth, keepaspectratio}
\begin{tabular}{l ccc}
\toprule
\textbf{Method} & Latency (ms) $\downarrow$ & GPU Mem (GB) $\downarrow$ & Throughput (samp/s) $\uparrow$ \\
\midrule
MCP & 0.45 & -- & 2206.40 \\
Perplexity & 0.41 & -- & 2417.03 \\
MTE & 0.45 & -- & 2238.20 \\
CCP & 0.50 & -- & 1998.38 \\
\midrule
SAPLMA & 5.10 & 21.71 & 196.20 \\
Factoscope & 4.69 & 32.52 & 213.36 \\
Lookback & 4.88 & 33.49 & 205.04 \\
UHead & 6.77 & 51.66 & 147.76 \\
\rowcolor{gray!30}
\textbf{\m} & 7.21 & 42.39 & 138.37 \\
\bottomrule
\end{tabular}
\end{adjustbox}
\caption{Inference efficiency on the HotpotQA test split.}
\label{tab:efficiency}
\end{table}

\begin{table}[t]
\centering
\setlength{\tabcolsep}{2pt}
\begin{adjustbox}{max width=\linewidth, keepaspectratio}
\begin{tabular}{l ccc}
\toprule
\textbf{Method} & Latency (ms) $\downarrow$ & GPU Mem (GB) $\downarrow$ & Throughput (samp/s) $\uparrow$ \\
\midrule
\multicolumn{4}{l}{\textbf{UHead}} \\
PS & 0.29 & 15.09 & 3450.62 \\
TS & 0.56 & 15.09 & 1788.41 \\
ISO & 0.33 & 15.09 & 3058.47 \\
BWH & 20.42 & 15.95 & 48.96 \\
\rowcolor{gray!30}
R-Log & 2.45 & 15.95 & 408.00 \\
\midrule
\multicolumn{4}{l}{\textbf{\m}} \\
PS & 0.30 & 15.09 & 3326.18 \\
TS & 0.55 & 15.09 & 1810.64 \\
ISO & 0.34 & 15.09 & 2945.73 \\
BWH & 19.34 & 15.10 & 51.72 \\
\rowcolor{gray!30}
\textbf{R-Log} & 4.90 & 15.10 & 204.00 \\
\bottomrule
\end{tabular}
\end{adjustbox}
\caption{Post-calibration efficiency on the HotpotQA test split.}
\label{tab:efficiency_post}
\end{table}

\section{Hardware and Efficiency}\label{sec:app_hardware_eff}
All experiments were conducted on a high-performance computing node running Red Hat Enterprise Linux 9.7, with one NVIDIA B200 GPU (180 GB VRAM), 12 CPU cores from an Intel Xeon Platinum 8570 processor, and 128 GB system memory. This hardware budget is sufficient to support the full offline workflow, including backbone generation, judge-based verification, cached-feature extraction, head training, and calibration evaluation.

Table~\ref{tab:efficiency_post} complements this base-head view with a calibrator-specific post-hoc efficiency accounting for UHead and \m under the five calibration methods discussed in Section~\ref{sec:evaluation}. The \m columns use the same best-performing uq head selected for Table~\ref{tab:efficiency}. Several trends are clear from the table. First, the scalar calibrators remain extremely cheap: PS and ISO stay around 0.3 ms per sample, while TS is only slightly slower at about 0.55 ms, so all three still support roughly 1.8k--3.5k samples/s. Second, the main separation is temporal rather than memory-related: GPU usage stays near 15 GB across almost all rows, whereas latency varies by nearly two orders of magnitude. Third, R-Log occupies a practical middle ground between scalar remapping and the heaviest structured alternative. It is substantially faster than BWH for both head families (408 vs.\ 48.96 samples/s on UHead; 204 vs.\ 51.72 samples/s on \m) while preserving the same structured-calibration regime, suggesting that reasoning-aware correction does not necessarily require paying the full binwise cost.

\section{Hyperparameter Space}

Table~\ref{tab:search-space} summarizes the representative selection space retained in the current implementation, with the final main-paper choices highlighted in bold. On the head side, the final configuration uses width \textbf{512}, a \textbf{two-layer} scoped encoder with \textbf{8} attention heads, dropout \textbf{0.1}, \textbf{balanced BCE}, and label smoothing \textbf{0.02}. These settings keep the first-stage predictor lightweight while preserving enough capacity for conclusion-focused refinement.

The frozen-feature recipe is likewise compact: hidden representations from $\boldsymbol{[-1,-2,-4,-8]}$ are fused with weights \textbf{[0.4, 0.3, 0.2, 0.1]}, together with \textbf{top-6} token-probability features and attention summaries from \textbf{layer $\boldsymbol{-1}$} only. On the optimization side, the final profile uses \textbf{30} epochs, learning rate $\boldsymbol{1.5 \times 10^{-4}}$, weight decay \textbf{0.05}, and warmup ratio \textbf{0.06}. For calibration fitting, the reported setup uses \textbf{auto} feature mode, tunes on the \textbf{validation} split, and enables the full dense-feature regime only once the tuning set reaches \textbf{1500} samples. The table should therefore be read as a concise representative search space rather than as an exhaustive Cartesian sweep over every possible configuration.
\begin{table}[t]
\centering
\footnotesize
\begin{adjustbox}{max width=\linewidth, keepaspectratio}
\begin{tabular}{p{0.28\linewidth}p{0.64\linewidth}}
\toprule
\textbf{Parameter} & \textbf{Values (bold = final)} \\
\midrule
\multicolumn{2}{l}{\textit{Head parameters}} \\
Head width $d_h$ & 256, \textbf{512}, 640 \\
Encoder layers & 1, \textbf{2}, 3 \\
Attention heads & 4, \textbf{8} \\
Dropout & 0.0, \textbf{0.1}, 0.2 \\
Loss type & BCE, \textbf{balanced BCE}, focal \\
Label smoothing & 0.00, \textbf{0.02}, 0.05 \\
\midrule
\multicolumn{2}{l}{\textit{Feature parameters}} \\
Hidden layers & $[-1,-2]$, $\boldsymbol{[-1,-2,-4,-8]}$, all \\
Fusion weights & uniform, \textbf{[0.4, 0.3, 0.2, 0.1]} \\
Top-$k$ token & 4, \textbf{6}, 8 \\
Attention layers & none, $\boldsymbol{-1}$, $[-1,-2]$ \\
\midrule
\multicolumn{2}{l}{\textit{Training / calibration parameters}} \\
Epochs & 20, \textbf{30}, 40 \\
Learning rate & $1.0\times10^{-4}$, $\boldsymbol{1.5 \times 10^{-4}}$, $2.0\times10^{-4}$ \\
Weight decay & 0.01, \textbf{0.05}, 0.10 \\
Warmup ratio & 0.00, \textbf{0.06}, 0.10 \\
Feature mode & compact, \textbf{auto}, full \\
Min samples & 1000, \textbf{1500}, 2000 \\
Tune split & \textbf{validation} \\
\bottomrule
\end{tabular}
\end{adjustbox}
\caption{Representative hyperparameter search space. \textbf{Bold} values denote the final configuration.}
\label{tab:search-space}
\end{table}

\begin{table*}[bt]
\centering
\setlength{\tabcolsep}{2pt}
\begin{adjustbox}{max width=\textwidth, keepaspectratio, max height=0.8\textheight}
\begin{tabular}{l ccc ccc ccc ccc}
\toprule
& \multicolumn{3}{c}{\textbf{HotpotQA (ID)}} & \multicolumn{3}{c}{\textbf{MuSiQue (OOD)}} & \multicolumn{3}{c}{\textbf{StrategyQA (OOD)}} & \multicolumn{3}{c}{\textbf{bAbI (OOD)}} \\
\cmidrule(lr){2-4} \cmidrule(lr){5-7} \cmidrule(lr){8-10} \cmidrule(lr){11-13}
\textbf{Method} & Acc$\uparrow$ & AUROC$\uparrow$ & PRAUC$\uparrow$ & Acc$\uparrow$ & AUROC$\uparrow$ & PRAUC$\uparrow$ & Acc$\uparrow$ & AUROC$\uparrow$ & PRAUC$\uparrow$ & Acc$\uparrow$ & AUROC$\uparrow$ & PRAUC$\uparrow$ \\
\midrule
Random & 0.491 & 0.483 & 0.525 & 0.496 & 0.481 & 0.257 & 0.459 & 0.472 & 0.698 & 0.506 & 0.504 & 0.559 \\
MCP & 0.566 & 0.548 & 0.603 & 0.522 & 0.593 & 0.324 & 0.703 & 0.618 & 0.761 & \underline{0.569} & 0.529 & 0.556 \\
Perplexity & 0.498 & 0.510 & 0.568 & 0.520 & 0.489 & 0.262 & 0.459 & 0.506 & 0.781 & 0.497 & 0.517 & 0.574 \\
MTE & 0.570 & 0.547 & 0.559 & 0.529 & 0.589 & 0.310 & 0.635 & 0.581 & 0.742 & 0.549 & 0.538 & 0.568 \\
CCP & 0.501 & 0.514 & 0.571 & \underline{0.538} & 0.493 & 0.269 & 0.463 & 0.511 & 0.786 & 0.501 & 0.522 & \underline{0.579} \\
\midrule
SAPLMA & 0.628 & \underline{0.707} & \underline{0.726} & 0.449 & 0.557 & 0.307 & 0.603 & 0.539 & 0.752 & 0.536 & 0.519 & 0.560 \\
Factoscope & 0.613 & 0.692 & 0.701 & 0.297 & 0.572 & 0.321 & \underline{0.760} & 0.631 & 0.806 & 0.546 & \underline{0.549} & 0.552 \\
Lookback & \underline{0.640} & 0.695 & 0.702 & 0.462 & \textbf{0.615} & \textbf{0.360} & 0.747 & \underline{0.675} & 0.825 & 0.555 & 0.527 & 0.534 \\
UHead & 0.618 & 0.698 & 0.711 & 0.480 & \underline{0.610} & 0.321 & 0.655 & 0.659 & \underline{0.875} & 0.544 & 0.528 & 0.554 \\
\rowcolor{gray!30}
\textbf{\m} & \textbf{0.651} & \textbf{0.719} & \textbf{0.727} & \textbf{0.626} & 0.607 & \underline{0.351} & \textbf{0.784} & \textbf{0.742} & \textbf{0.889} & \textbf{0.582} & \textbf{0.587} & \textbf{0.589} \\
\bottomrule
\end{tabular}
\end{adjustbox}
\caption{Mistral-7B-Instruct-v0.3 discrimination results on four datasets.}
\label{tab:result_2}
\end{table*}

\begin{table*}[t]
\centering
\setlength{\tabcolsep}{4pt}
\begin{adjustbox}{max width=\textwidth, keepaspectratio}
\begin{tabular}{l ccc ccc ccc ccc}
\toprule
& \multicolumn{3}{c}{\textbf{HotpotQA (ID)}} & \multicolumn{3}{c}{\textbf{MuSiQue (OOD)}} & \multicolumn{3}{c}{\textbf{StrategyQA (OOD)}} & \multicolumn{3}{c}{\textbf{bAbI (OOD)}} \\
\cmidrule(lr){2-4} \cmidrule(lr){5-7} \cmidrule(lr){8-10} \cmidrule(lr){11-13}
\textbf{Method} & ECE$\downarrow$ & Brier$\downarrow$ & Risk@90$\downarrow$ & ECE$\downarrow$ & Brier$\downarrow$ & Risk@90$\downarrow$ & ECE$\downarrow$ & Brier$\downarrow$ & Risk@90$\downarrow$ & ECE$\downarrow$ & Brier$\downarrow$ & Risk@90$\downarrow$ \\
\midrule
PS & 0.023 & 0.213 & 0.325 & \textbf{0.065} & 0.187 & 0.233 & 0.280 & 0.285 & 0.310 & 0.042 & 0.243 & 0.393 \\
TS & \underline{0.017} & 0.216 & 0.332 & \underline{0.069} & 0.200 & 0.270 & 0.290 & 0.295 & 0.320 & \textbf{0.031} & 0.246 & 0.420 \\
ISO & 0.026 & 0.214 & 0.327 & 0.075 & 0.187 & 0.245 & 0.285 & \underline{0.280} & \underline{0.305} & \underline{0.039} & 0.242 & 0.389 \\
BWH & 0.045 & \underline{0.047} & \underline{0.029} & 0.141 & \textbf{0.126} & \textbf{0.097} & \underline{0.275} & 0.290 & 0.315 & 0.052 & \underline{0.220} & \textbf{0.261} \\
\rowcolor{gray!30}
\textbf{R-Log} & \textbf{0.014} & \textbf{0.044} & \textbf{0.025} & 0.072 & \underline{0.144} & \underline{0.194} & \textbf{0.270} & \textbf{0.270} & \textbf{0.299} & 0.052 & \textbf{0.210} & \underline{0.307} \\
\bottomrule
\end{tabular}
\end{adjustbox}
\caption{Mistral-7B-Instruct-v0.3 calibration results on four datasets based on \m.}
\label{tab:calibration_2}
\end{table*}

\begin{table*}[t]
\centering
\setlength{\tabcolsep}{2pt}
\begin{adjustbox}{max width=\textwidth, keepaspectratio, max height=0.8\textheight}
\begin{tabular}{l ccc ccc ccc ccc}
\toprule
& \multicolumn{3}{c}{\textbf{HotpotQA (ID)}} & \multicolumn{3}{c}{\textbf{MuSiQue (OOD)}} & \multicolumn{3}{c}{\textbf{StrategyQA (OOD)}} & \multicolumn{3}{c}{\textbf{bAbI (OOD)}} \\
\cmidrule(lr){2-4} \cmidrule(lr){5-7} \cmidrule(lr){8-10} \cmidrule(lr){11-13}
\textbf{Method} & Acc$\uparrow$ & AUROC$\uparrow$ & PRAUC$\uparrow$ & Acc$\uparrow$ & AUROC$\uparrow$ & PRAUC$\uparrow$ & Acc$\uparrow$ & AUROC$\uparrow$ & PRAUC$\uparrow$ & Acc$\uparrow$ & AUROC$\uparrow$ & PRAUC$\uparrow$ \\
\midrule
Random & 0.499 & 0.493 & 0.752 & 0.501 & 0.482 & 0.391 & 0.486 & 0.462 & 0.906 & 0.507 & 0.511 & 0.739 \\
MCP & 0.579 & 0.538 & 0.788 & 0.477 & 0.499 & 0.377 & 0.671 & 0.549 & 0.921 & 0.717 & 0.503 & 0.676 \\
Perplexity & 0.510 & 0.448 & 0.736 & \underline{0.540} & 0.522 & 0.406 & 0.760 & 0.489 & 0.901 & 0.698 & 0.524 & 0.709 \\
MTE & 0.535 & 0.534 & 0.784 & 0.515 & 0.501 & 0.382 & 0.457 & 0.539 & 0.918 & 0.707 & 0.534 & 0.724 \\
CCP & 0.496 & 0.451 & 0.735 & 0.538 & 0.525 & 0.409 & 0.758 & 0.492 & 0.904 & 0.701 & 0.528 & 0.713 \\
\midrule
SAPLMA & 0.780 & 0.720 & 0.851 & \textbf{0.545} & \textbf{0.647} & \underline{0.565} & 0.884 & 0.654 & 0.924 & 0.716 & 0.701 & 0.842 \\
Factoscope & 0.775 & \textbf{0.749} & \underline{0.872} & 0.539 & 0.630 & 0.530 & \underline{0.887} & 0.627 & 0.922 & 0.732 & 0.729 & 0.856 \\
Lookback & 0.771 & 0.736 & 0.865 & 0.473 & 0.609 & 0.517 & \underline{0.887} & 0.589 & 0.917 & \textbf{0.751} & \underline{0.745} & \underline{0.862} \\
UHead & \underline{0.786} & 0.740 & 0.870 & 0.538 & 0.635 & \textbf{0.570} & \underline{0.887} & \underline{0.669} & \underline{0.927} & 0.728 & 0.736 & 0.861 \\
\rowcolor{gray!30}
\textbf{\m} & \textbf{0.794} & \underline{0.746} & \textbf{0.886} & 0.533 & \underline{0.640} & 0.562 & \textbf{0.914} & \textbf{0.690} & \textbf{0.954} & \underline{0.737} & \textbf{0.759} & \textbf{0.888} \\
\bottomrule
\end{tabular}
\end{adjustbox}
\caption{Gemma-2-9B-it discrimination results on four datasets. }
\label{tab:result_3}
\end{table*}

\begin{table*}[t]
\centering
\setlength{\tabcolsep}{4pt}
\begin{adjustbox}{max width=\textwidth, keepaspectratio}
\begin{tabular}{l ccc ccc ccc ccc}
\toprule
& \multicolumn{3}{c}{\textbf{HotpotQA (ID)}} & \multicolumn{3}{c}{\textbf{MuSiQue (OOD)}} & \multicolumn{3}{c}{\textbf{StrategyQA (OOD)}} & \multicolumn{3}{c}{\textbf{bAbI (OOD)}} \\
\cmidrule(lr){2-4} \cmidrule(lr){5-7} \cmidrule(lr){8-10} \cmidrule(lr){11-13}
\textbf{Method} & ECE$\downarrow$ & Brier$\downarrow$ & Risk@90$\downarrow$ & ECE$\downarrow$ & Brier$\downarrow$ & Risk@90$\downarrow$ & ECE$\downarrow$ & Brier$\downarrow$ & Risk@90$\downarrow$ & ECE$\downarrow$ & Brier$\downarrow$ & Risk@90$\downarrow$ \\
\midrule
PS & \underline{0.010} & 0.152 & \underline{0.182} & \underline{0.033} & 0.228 & 0.346 & 0.048 & 0.078 & \underline{0.074} & 0.049 & 0.168 & 0.231 \\
TS & 0.047 & 0.211 & 0.297 & 0.043 & 0.227 & 0.343 & \underline{0.033} & 0.101 & 0.086 & 0.041 & 0.183 & 0.225 \\
ISO & 0.021 & \underline{0.150} & 0.185 & 0.039 & 0.229 & 0.344 & \textbf{0.018} & \underline{0.076} & \underline{0.074} & \underline{0.034} & 0.162 & 0.221 \\
BWH & 0.189 & 0.202 & 0.200 & 0.061 & \textbf{0.156} & \textbf{0.217} & 0.048 & 0.078 & \underline{0.074} & 0.133 & \underline{0.142} & \textbf{0.125} \\
\rowcolor{gray!30}
\textbf{R-Log} & \textbf{0.009} & \textbf{0.034} & \textbf{0.025} & \textbf{0.027} & \underline{0.173} & \underline{0.240} & 0.041 & \textbf{0.071} & \textbf{0.055} & \textbf{0.031} & \textbf{0.126} & \underline{0.153} \\
\bottomrule
\end{tabular}
\end{adjustbox}
\caption{Gemma-2-9B-it calibration results on four datasets based on \m.}
\label{tab:calibration_3}
\end{table*}

\begin{table*}[t]
\centering
\setlength{\tabcolsep}{2pt}
\begin{adjustbox}{max width=\textwidth, keepaspectratio, max height=0.8\textheight}
\begin{tabular}{l ccc ccc ccc ccc}
\toprule
& \multicolumn{3}{c}{\textbf{HotpotQA (ID)}} & \multicolumn{3}{c}{\textbf{MuSiQue (OOD)}} & \multicolumn{3}{c}{\textbf{StrategyQA (OOD)}} & \multicolumn{3}{c}{\textbf{bAbI (OOD)}} \\
\cmidrule(lr){2-4} \cmidrule(lr){5-7} \cmidrule(lr){8-10} \cmidrule(lr){11-13}
\textbf{Method} & Acc$\uparrow$ & AUROC$\uparrow$ & PRAUC$\uparrow$ & Acc$\uparrow$ & AUROC$\uparrow$ & PRAUC$\uparrow$ & Acc$\uparrow$ & AUROC$\uparrow$ & PRAUC$\uparrow$ & Acc$\uparrow$ & AUROC$\uparrow$ & PRAUC$\uparrow$ \\
\midrule
Random & 0.497 & 0.496 & 0.745 & 0.486 & 0.489 & 0.409 & 0.511 & 0.509 & 0.943 & 0.499 & 0.499 & 0.815 \\
MCP & 0.679 & 0.573 & 0.800 & 0.531 & 0.603 & 0.510 & \underline{0.936} & 0.542 & \underline{0.970} & 0.761 & 0.583 & \textbf{0.901} \\
Perplexity & 0.424 & 0.419 & 0.691 & 0.440 & 0.437 & 0.359 & 0.443 & 0.365 & 0.898 & 0.303 & 0.392 & 0.754 \\
MTE & 0.673 & 0.568 & 0.765 & 0.543 & 0.604 & 0.472 & \underline{0.936} & 0.635 & 0.961 & 0.756 & 0.599 & 0.861 \\
CCP & 0.414 & 0.423 & 0.694 & 0.441 & 0.435 & 0.358 & 0.083 & 0.445 & 0.840 & 0.307 & 0.396 & 0.756 \\
\midrule
SAPLMA & 0.787 & 0.765 & 0.871 & 0.603 & 0.691 & 0.606 & 0.902 & 0.649 & 0.932 & 0.781 & 0.649 & 0.871 \\
Factoscope & \underline{0.791} & 0.769 & 0.869 & 0.614 & \underline{0.720} & 0.614 & 0.912 & \underline{0.689} & 0.934 & 0.776 & 0.634 & 0.850 \\
Lookback & \underline{0.791} & 0.770 & 0.874 & 0.628 & 0.706 & 0.603 & 0.909 & 0.611 & 0.921 & \underline{0.788} & \underline{0.655} & 0.862 \\
UHead & 0.786 & \underline{0.775} & \underline{0.880} & \textbf{0.644} & \textbf{0.722} & \textbf{0.633} & 0.909 & 0.449 & 0.905 & 0.779 & 0.636 & 0.863 \\
\rowcolor{gray!30}
\textbf{\m} & \textbf{0.807} & \textbf{0.786} & \textbf{0.899} & \underline{0.640} & 0.714 & \underline{0.627} & \textbf{0.942} & \textbf{0.724} & \textbf{0.971} & \textbf{0.812} & \textbf{0.673} & \underline{0.893} \\
\bottomrule
\end{tabular}
\end{adjustbox}
\caption{Phi-4 discrimination results on four datasets. }
\label{tab:result_4}
\end{table*}

\begin{table*}[t]
\centering
\setlength{\tabcolsep}{4pt}
\begin{adjustbox}{max width=\textwidth, keepaspectratio}
\begin{tabular}{l ccc ccc ccc ccc}
\toprule
& \multicolumn{3}{c}{\textbf{HotpotQA (ID)}} & \multicolumn{3}{c}{\textbf{MuSiQue (OOD)}} & \multicolumn{3}{c}{\textbf{StrategyQA (OOD)}} & \multicolumn{3}{c}{\textbf{bAbI (OOD)}} \\
\cmidrule(lr){2-4} \cmidrule(lr){5-7} \cmidrule(lr){8-10} \cmidrule(lr){11-13}
\textbf{Method} & ECE$\downarrow$ & Brier$\downarrow$ & Risk@90$\downarrow$ & ECE$\downarrow$ & Brier$\downarrow$ & Risk@90$\downarrow$ & ECE$\downarrow$ & Brier$\downarrow$ & Risk@90$\downarrow$ & ECE$\downarrow$ & Brier$\downarrow$ & Risk@90$\downarrow$ \\
\midrule
PS & 0.027 & 0.143 & 0.165 & 0.057 & 0.213 & 0.326 & \underline{0.025} & 0.055 & \underline{0.048} & 0.027 & 0.145 & 0.163 \\
TS & 0.014 & 0.179 & 0.240 & \underline{0.042} & 0.213 & 0.325 & 0.034 & 0.058 & 0.050 & 0.112 & 0.226 & 0.312 \\
ISO & \underline{0.008} & \underline{0.141} & \underline{0.162} & 0.047 & 0.212 & 0.335 & 0.029 & \underline{0.054} & \underline{0.048} & \underline{0.019} & 0.144 & 0.164 \\
BWH & 0.153 & 0.172 & 0.171 & 0.061 & \textbf{0.132} & \textbf{0.143} & \underline{0.025} & 0.055 & \underline{0.048} & 0.088 & \underline{0.098} & \textbf{0.080} \\
\rowcolor{gray!30}
\textbf{R-Log} & \textbf{0.007} & \textbf{0.036} & \textbf{0.023} & \textbf{0.035} & \underline{0.176} & \underline{0.247} & \textbf{0.024} & \textbf{0.052} & \textbf{0.040} & \textbf{0.017} & \textbf{0.090} & \underline{0.085} \\
\bottomrule
\end{tabular}
\end{adjustbox}
\caption{Phi-4 calibration results on four datasets based on \m.}
\label{tab:calibration_4}
\end{table*}

\section{Additional Evaluation Results}
The main paper reports the Llama-based results in the body, and this appendix extends the same evaluation protocol to Mistral-7B-Instruct-v0.3, Gemma-2-9B-it, and Phi-4. Unless otherwise stated, these runs follow the representative search space summarized in Table~\ref{tab:search-space}, with the \textbf{bold} entries serving as the default configuration reused across backbones rather than re-tuning a separate head for every appendix table.

Specifically, the cross-backbone performance tables are Table~\ref{tab:result_2}, Table~\ref{tab:result_3}, and Table~\ref{tab:result_4}; the corresponding calibration tables are Table~\ref{tab:calibration_2}, Table~\ref{tab:calibration_3}, and Table~\ref{tab:calibration_4}. Across these additional backbones, the same qualitative pattern remains visible: \m is typically strongest or near-strongest on response-level discrimination, with especially stable gains on HotpotQA and StrategyQA, while remaining competitive on the harder MuSiQue OOD dataset. The calibration tables show a similar trend for calibration reliability: reasoning consistency-aware fitting continues to reduce ECE, Brier score, and often Risk@90 relative to simpler scalar calibrators across most settings, although MuSiQue remains the most difficult OOD target. Taken together, these appendix tables indicate that the method's gains are not tied to a single backbone or a separately tuned appendix-only setup, but persist under the same lightweight design budget summarized by Table~\ref{tab:search-space}.

\section{Documentation Of Artifacts}

All artifacts used in this work are public research artifacts. Appendix~\ref{sec:app_data} documents the dataset splits and preprocessing, Appendix~\ref{sec:app_model} summarizes the backbone and judge models, Appendix~\ref{sec:app_baseline} documents the compared baseline families, and Appendix~\ref{sec:app_prompt_templates} provides the generation and judge prompt templates that define the derived cached artifacts consumed by the pipeline.

The dataset coverage is entirely English and spans four complementary reasoning settings: HotpotQA and MuSiQue are open-domain multi-hop question answering benchmarks over natural text, StrategyQA is a yes/no commonsense reasoning benchmark over short fact lists, and bAbI is a synthetic controlled benchmark for state tracking and compositional reasoning. Accordingly, the main linguistic and reasoning phenomena covered by our artifacts include multi-document evidence aggregation, compositional multi-hop inference, implicit commonsense reasoning, entity/state tracking, and short-answer or yes/no answer generation.

These artifacts are task-oriented reasoning benchmarks rather than demographic surveys. They are not designed to represent particular demographic groups, and we do not use demographic attributes as supervision targets, evaluation datasets, or calibration features. Our derived caches preserve the original task scope at the response and claim levels so that the experimental setting remains aligned with the source benchmark documentation.

\section{GenAI Disclosure}
In the preparation of this work, the authors utilized Generative AI tools solely for the purpose of language refinement and improving readability. No AI tools were used to generate scientific concepts, experimental results, or the intellectual content of this paper. The authors have reviewed all AI-assisted edits and take full responsibility for the final content of the manuscript.

\clearpage
\onecolumn
\newlength{\appendixcardwidth}
\setlength{\appendixcardwidth}{\textwidth}
\newcommand{\cardtitle}[2]{%
  \vspace*{2pt}%
  \noindent\textbf{\textcolor{#1}{#2}}\par
  \vspace{1pt}%
  \noindent\textcolor{#1}{\rule{\linewidth}{0.5pt}}\par
  \vspace{4pt}%
}
\newcommand{\scoretag}[3][4.2em]{%
  \begingroup
  \setlength{\fboxsep}{0pt}%
  \colorbox{#2}{\parbox[c][1.55em][c]{#1}{\centering\footnotesize #3}}%
  \endgroup
}

\section{Prompt Templates}\label{sec:app_prompt_templates}

We use one generation prompt family and two judge prompt families. The generation side combines a dataset-specific instruction prompt with a shared output-format contract so that the target LLM emits short atomic reasoning steps followed by one conclusion line. The judge side is split into two prompts because conclusion correctness and step-level reasoning consistency supervise different cache fields and should expose different information to the verifier.

Throughout this section, we present the structural form rather than every dataset-specific wording variant. In the implementation, each dataset instantiates the same skeleton with task-specific question/context fields, answer-space constraints, and a small number of extra rules.

\subsection{Generation-side templates}

\noindent\makebox[\textwidth][c]{\fcolorbox{blue!60!black}{blue!4}{
\begin{minipage}[t]{\dimexpr\appendixcardwidth-2\fboxsep-2\fboxrule\relax}
\small
\renewcommand{\arraystretch}{1.22}
\cardtitle{blue!60!black}{Generation prompt skeleton}
\begin{tabular}{@{}p{0.20\linewidth}p{0.74\linewidth}@{}}
\textbf{Role / Goal} &
Target backbone LLM; solve the task with minimal evidence-based reasoning while emitting a stable reasoning/conclusion format.\\[4pt]
\textbf{Core template} &
\colorbox{blue!12}{\parbox{0.92\linewidth}{
\texttt{You are a careful reasoning assistant.}\\
\texttt{Task objective: \textcolor{blue!70!black}{<DATASET OBJECTIVE>}}\\[2pt]
\texttt{Output schema (must follow exactly):}\\
\texttt{Reasoning:}\\
\texttt{Step 1: <one atomic fact>}\\
\texttt{Step 2: <one atomic fact>}\\
\texttt{...}\\
\texttt{Conclusion: \textcolor{blue!70!black}{<DATASET-SPECIFIC CONCLUSION FORMAT>}}\\[2pt]
\texttt{Formatting constraints: plain text only; use only as many steps as needed;}\\
\texttt{keep each step short; no text before Reasoning or after Conclusion.}
}}\\[4pt]
\textbf{Dataset-specific fields} &
\textcolor{blue!70!black}{<DATASET OBJECTIVE>} changes with the task (e.g., multi-hop QA, yes/no reasoning, or controlled symbolic QA), and the conclusion schema is specialized to the required response type, e.g., a free-form entity, \texttt{yes/no}, a label ID, or a constrained option string.\\[4pt]
\textbf{Design intent} &
This shared system prompt standardizes the \texttt{Reasoning}/\texttt{Conclusion} split before claim extraction, while still allowing each dataset to impose its own answer-space constraints.
\end{tabular}
\end{minipage}}}

\vspace{8pt}
\noindent\makebox[\textwidth][c]{\fcolorbox{teal!60!black}{teal!4}{
\begin{minipage}[t]{\dimexpr\appendixcardwidth-2\fboxsep-2\fboxrule\relax}
\small
\renewcommand{\arraystretch}{1.22}
\cardtitle{teal!60!black}{Shared output contract}
\begin{tabular}{@{}p{0.20\linewidth}p{0.74\linewidth}@{}}
\textbf{Purpose} &
A shared contract is injected into every prompt to make downstream claim parsing and conclusion alignment robust across datasets.\\[4pt]
\textbf{Injected contract} &
\colorbox{teal!12}{\parbox{0.92\linewidth}{
\texttt{STRICT OUTPUT CONTRACT:}\\
\texttt{Objective: solve the task using minimal evidence-based reasoning.}\\
\texttt{Output schema (exact):}\\
\texttt{Reasoning:}\\
\texttt{Step 1: <one atomic fact>}\\
\texttt{Step 2: <one atomic fact>}\\
\texttt{...}\\
\texttt{Conclusion: <dataset-specific final conclusion>}\\[2pt]
\texttt{Rules: plain text only; 1--8 Step lines;}\\
\texttt{the Conclusion line contains only the final conclusion;}\\
\texttt{no text before `Reasoning:' or after `Conclusion:'.}
}}\\[4pt]
\textbf{Operational effect} &
Because the same contract is attached to each dataset prompt, the reasoning block can be segmented into atomic claims with a uniform parser, and the final conclusion span can be isolated as the only region consumed by the alignment-aware lightweight UQ module.\\[4pt]
\textbf{Typical user fields} &
The user message then instantiates the actual sample via question text and optional context/passages/options, e.g., \textcolor{teal!70!black}{<QUESTION>}, \textcolor{teal!70!black}{<CONTEXT OR PASSAGES>}, and dataset-specific option labels if the answer space is closed.
\end{tabular}
\end{minipage}}}

\subsection{Judge-side templates}

\noindent\makebox[\textwidth][c]{\fcolorbox{purple!65!black}{purple!4}{
\begin{minipage}[t]{\dimexpr\appendixcardwidth-2\fboxsep-2\fboxrule\relax}
\small
\renewcommand{\arraystretch}{1.22}
\cardtitle{purple!65!black}{Stage 1: conclusion verification}
\begin{tabular}{@{}p{0.20\linewidth}p{0.74\linewidth}@{}}
\textbf{Role / Goal} &
Judge model; decide whether the model's \emph{conclusion} matches the ground-truth conclusion. This stage produces the response-level correctness label and conclusion verdict.\\[4pt]
\textbf{Prompt skeleton} &
\colorbox{purple!12}{\parbox{0.92\linewidth}{
\texttt{You are an exact conclusion verifier.}\\
\texttt{Objective: decide whether the model's conclusion matches the ground-truth conclusion.}\\
\texttt{Input fields:}\\
\texttt{Ground Truth Conclusion: \textcolor{purple!70!black}{<REFERENCE CONCLUSION>}}\\
\texttt{Conclusion Claim: \textcolor{purple!70!black}{<MODEL CONCLUSION>}}\\
\texttt{Output schema (strict JSON): \{"correct": true/false\}}\\[2pt]
\texttt{Optional label set: [\textcolor{purple!70!black}{<ALLOWED LABELS>}]}\\
\texttt{Normalize to one allowed label when the task uses a closed response space.}
}}\\[4pt]
\textbf{Returned structure} &
The minimal schema is \texttt{\{"correct": true/false\}}. When guided analysis is enabled, the judge may instead return \texttt{\{"analysis": "...", "correct": true/false\}}, but the correctness field remains mandatory and machine-checked.\\[4pt]
\textbf{Design intent} &
Stage 1 sees the reference conclusion because it is responsible only for \emph{response correctness}. It does \emph{not} label internal reasoning steps; it only supervises the response-level target used in training and evaluation.
\end{tabular}
\end{minipage}}}

\vspace{8pt}
\noindent\makebox[\textwidth][c]{\fcolorbox{violet!65!black}{violet!4}{
\begin{minipage}[t]{\dimexpr\appendixcardwidth-2\fboxsep-2\fboxrule\relax}
\small
\renewcommand{\arraystretch}{1.22}
\cardtitle{violet!65!black}{Stage 2: reasoning consistency}
\begin{tabular}{@{}p{0.20\linewidth}p{0.74\linewidth}@{}}
\textbf{Role / Goal} &
Judge model; verify each extracted reasoning step for consistency with the question, the provided context/passages, and the model's own conclusion. This stage produces the step-level labels used only by the reasoning consistency-aware calibration stage.\\[4pt]
\textbf{Prompt skeleton} &
\colorbox{violet!12}{\parbox{0.92\linewidth}{
\texttt{You are an exact reasoning-step verifier.}\\
\texttt{Objective: evaluate each reasoning step for contextual support and logical consistency toward the conclusion.}\\
\texttt{Input fields:}\\
\texttt{Question: \textcolor{violet!70!black}{<QUESTION>}}\\
\texttt{Context / Passages: \textcolor{violet!70!black}{<CONTEXT OR PASSAGES>}}\\
\texttt{Reasoning:}\\
\texttt{Step 1: \textcolor{violet!70!black}{<CLAIM 1>}}\\
\texttt{Step 2: \textcolor{violet!70!black}{<CLAIM 2>}}\\
\texttt{...}\\
\texttt{Conclusion: \textcolor{violet!70!black}{<MODEL CONCLUSION>}}\\
\texttt{Output schema (strict JSON): \{"reasoning\_verified": [0,1,...]\}}
}}\\[4pt]
\textbf{Returned structure} &
The returned array length must match the number of extracted reasoning steps exactly. With guided analysis enabled, the judge may prepend a short explanation field, but the required machine-readable payload remains \texttt{\{"reasoning\_verified": [0,1,...]\}}.\\[4pt]
\textbf{Leakage control} &
Unlike Stage 1, this prompt intentionally \emph{omits the ground-truth conclusion}. The judge only inspects the question, the provided context/passages, the extracted reasoning steps, and the conclusion claim. This prevents step-level labels from trivially encoding the gold conclusion and keeps them as residual reasoning-consistency evidence rather than a second direct supervision signal for the predictor.
\end{tabular}
\end{minipage}}}

\vspace{8pt}
\noindent\makebox[\textwidth][c]{\fcolorbox{gray!60}{gray!6}{
\begin{minipage}[t]{\dimexpr\appendixcardwidth-2\fboxsep-2\fboxrule\relax}
\footnotesize\raggedright
\textbf{Implementation notes.}
\begin{itemize}[leftmargin=5mm,itemsep=2pt,parsep=0pt]
    \item Generation prompts may be truncated to a fixed token budget, but the \texttt{Reasoning}/\texttt{Conclusion} contract is preserved.
    \item Judge prompts use strict JSON schemas to minimize parsing failures; arrays with the wrong length or values outside \{0,1\} are rejected.
    \item For closed-label datasets such as spatial or multiple-choice tasks, Stage-1 prompts optionally provide an explicit allowed-label set to force normalized verdicts.
    \item Open-ended conclusion grading reuses the same Stage-1 structure but compares the model conclusion against a free-form reference conclusion rather than exact label IDs.
\end{itemize}
\end{minipage}}}

\clearpage

\section{Case Studies}\label{sec:app_case_studies}

We provide six representative examples from the main Llama run, covering both the source benchmark and transfer benchmarks. Examples A and B illustrate two characteristic reasoning consistency-aware calibration behaviors on HotpotQA: structured reasoning signals can raise confidence for a correct but under-confident response, and they can also down-weight a locally plausible but question-mismatched conclusion. Examples C--F extend the same lens to OOD settings, where confidence increases when transferred reasoning remains coherent and decreases when multi-hop composition drifts away from the target relation.

\noindent\makebox[\textwidth][c]{\fcolorbox{blue!60!black}{blue!4}{
\begin{minipage}[t]{\dimexpr\appendixcardwidth-2\fboxsep-2\fboxrule\relax}
\small
\renewcommand{\arraystretch}{1.22}
\cardtitle{blue!60!black}{Example A: correct response rescued}
\begin{tabular}{@{}p{0.18\linewidth}p{0.76\linewidth}@{}}
\textbf{Question} &
Does \textcolor{blue!70!black}{Dashboard Confessional} have more members than \textcolor{blue!70!black}{World Party}?\\[4pt]
\textbf{LLM output} &
\parbox[t]{0.92\linewidth}{\footnotesize\raggedright
Reasoning: \textbf{Step 1:} Dashboard Confessional is an American emo band led by singer-songwriter Chris Carrabba. \textbf{Step 2:} World Party is a British alternative rock band, essentially the solo project of Karl Wallinger. \textbf{Step 3:} Dashboard Confessional includes Chris Carrabba, Mike Marsh, and Jay Orpin. \textbf{Step 4:} World Party includes only Karl Wallinger. Conclusion: \textcolor{green!50!black}{Yes}. Dashboard Confessional has more members than World Party.}\\[4pt]
\textbf{Judge / scores} &
\parbox[t]{0.92\linewidth}{%
\scoretag[7.2em]{green!18}{correct (1)} \quad
\scoretag{green!18}{S1: 1} \scoretag{green!18}{S2: 1} \scoretag{green!18}{S3: 1} \scoretag{green!18}{S4: 1}\\[4pt]
Alignment-aware UQ: 0.089 \quad Calibrated confidence: \scoretag{green!18}{0.998}}\\[5pt]
\textbf{Interpretation} &
The alignment-aware lightweight UQ stage remains cautious, but the reasoning trace forms a clean \textcolor{blue!70!black}{entity-count comparison} with fully supported intermediate steps. The reasoning consistency-aware calibrator therefore upgrades the final confidence sharply.
\end{tabular}
\end{minipage}}}

\vspace{8pt}
\noindent\makebox[\textwidth][c]{\fcolorbox{red!65!black}{red!4}{
\begin{minipage}[t]{\dimexpr\appendixcardwidth-2\fboxsep-2\fboxrule\relax}
\small
\renewcommand{\arraystretch}{1.22}
\cardtitle{red!65!black}{Example B: wrong abstraction suppressed}
\begin{tabular}{@{}p{0.18\linewidth}p{0.76\linewidth}@{}}
\textbf{Question} &
Wexner Graduate Fellowships are given to students who show an ability of an individual to ``lead'' or guide who?\\[4pt]
\textbf{LLM output} &
\parbox[t]{0.92\linewidth}{\footnotesize\raggedright
Reasoning: \textbf{Step 1:} The Wexner Graduate Fellowship Program supports graduate students planning a career related to Judaism. \textbf{Step 2:} The program selects students preparing for careers in the rabbinate, cantorate, academic Jewish studies, and Jewish communal service. \textbf{Step 3:} Leadership is a skill encompassing the ability to ``lead'' or guide other individuals, teams, or entire organizations. \textbf{Step 4:} The asked ability is therefore \textcolor{red!70!black}{leadership}. Conclusion: \textcolor{red!70!black}{leadership}.}\\[4pt]
\textbf{Judge / scores} &
\parbox[t]{0.92\linewidth}{%
\scoretag[7.2em]{red!18}{incorrect (0)} \quad
\scoretag{green!18}{S1: 1} \scoretag{green!18}{S2: 1} \scoretag{red!18}{S3: 0} \scoretag{red!18}{S4: 0}\\[4pt]
Alignment-aware UQ: 0.649 \quad Calibrated confidence: \scoretag{red!18}{0.096}}\\[5pt]
\textbf{Interpretation} &
The head is initially overconfident because the conclusion state looks locally plausible. However, the chain drifts from the fellowship description to a \textcolor{red!70!black}{dictionary-style abstraction} instead of answering the target argument of ``guide who.'' Reasoning consistency-aware calibration detects this mismatch and lowers confidence sharply.
\end{tabular}
\end{minipage}}}

\vspace{8pt}
\noindent\makebox[\textwidth][c]{\fcolorbox{blue!60!black}{blue!4}{
\begin{minipage}[t]{\dimexpr\appendixcardwidth-2\fboxsep-2\fboxrule\relax}
\small
\renewcommand{\arraystretch}{1.22}
\cardtitle{blue!60!black}{Example C: clean OOD chain reinforced}
\begin{tabular}{@{}p{0.18\linewidth}p{0.76\linewidth}@{}}
\textbf{Question} &
In what countries did \textcolor{blue!70!black}{Prettejohn} work at universities?\\[4pt]
\textbf{LLM output} &
\parbox[t]{0.92\linewidth}{\footnotesize\raggedright
Reasoning: \textbf{Step 1:} Elizabeth Prettejohn worked at the University of Bristol. \textbf{Step 2:} Elizabeth Prettejohn worked at the University of York. \textbf{Step 3:} The University of Bristol is located in England. \textbf{Step 4:} The University of York is located in England. Conclusion: \textcolor{green!50!black}{England}.}\\[4pt]
\textbf{Judge / scores} &
\parbox[t]{0.92\linewidth}{%
\scoretag[7.2em]{green!18}{correct (1)} \quad
\scoretag{green!18}{S1: 1} \scoretag{green!18}{S2: 1} \scoretag{green!18}{S3: 1} \scoretag{green!18}{S4: 1}\\[4pt]
Alignment-aware UQ: 0.325 \quad Calibrated confidence: \scoretag{green!18}{0.919}}\\[5pt]
\textbf{Interpretation} &
This transfer example shows that the same pattern persists under OOD evaluation. The alignment-aware lightweight UQ stage remains uncertain, but the reasoning chain forms a compact \textcolor{blue!70!black}{location-alignment argument}: both institutions map to the same country. Reasoning consistency-aware calibration recognizes that repeated agreement and raises confidence substantially.
\end{tabular}
\end{minipage}}}

\vspace{8pt}
\noindent\makebox[\textwidth][c]{\fcolorbox{red!65!black}{red!4}{
\begin{minipage}[t]{\dimexpr\appendixcardwidth-2\fboxsep-2\fboxrule\relax}
\small
\renewcommand{\arraystretch}{1.22}
\cardtitle{red!65!black}{Example D: final comparison fails}
\begin{tabular}{@{}p{0.18\linewidth}p{0.76\linewidth}@{}}
\textbf{Question} &
Which film has the director who was born first, \textcolor{red!70!black}{Horrible Bosses} or \textcolor{red!70!black}{Some Kind of Life}?\\[4pt]
\textbf{LLM output} &
\parbox[t]{0.92\linewidth}{\footnotesize\raggedright
Reasoning: \textbf{Step 1:} Seth Gordon directed \textit{Horrible Bosses}. \textbf{Step 2:} Seth Gordon was born in 1976. \textbf{Step 3:} Julian Jarrold directed \textit{Some Kind of Life}. \textbf{Step 4:} Julian Jarrold was born in 1960. \textbf{Step 5:} Since 1976 is earlier than 1960, the answer is \textcolor{red!70!black}{Horrible Bosses}. Conclusion: \textcolor{red!70!black}{Horrible Bosses}.}\\[4pt]
\textbf{Judge / scores} &
\parbox[t]{0.92\linewidth}{%
\scoretag[7.2em]{red!18}{incorrect (0)} \quad
\scoretag{green!18}{S1: 1} \scoretag{green!18}{S2: 1} \scoretag{green!18}{S3: 1} \scoretag{green!18}{S4: 1} \scoretag{red!18}{S5: 0}\\[4pt]
Alignment-aware UQ: 0.758 \quad Calibrated confidence: \scoretag{red!18}{0.242}}\\[5pt]
\textbf{Interpretation} &
This transfer case complements Example B with a different failure mode. The local entity links and birth-year facts are correct, but the final comparison step is wrong: 1960 is earlier than 1976, so the conclusion should be \textcolor{red!70!black}{Some Kind of Life}. Reasoning consistency-aware calibration therefore suppresses confidence once the chain makes an explicit inconsistent comparison instead of just trusting the locally plausible entities and years.
\end{tabular}
\end{minipage}}}

\vspace{8pt}
\noindent\makebox[\textwidth][c]{\fcolorbox{blue!60!black}{blue!4}{
\begin{minipage}[t]{\dimexpr\appendixcardwidth-2\fboxsep-2\fboxrule\relax}
\small
\renewcommand{\arraystretch}{1.22}
\cardtitle{blue!60!black}{Example E: short composition transfers}
\begin{tabular}{@{}p{0.18\linewidth}p{0.76\linewidth}@{}}
\textbf{Question} &
Who is the spouse of the director of film \textcolor{blue!70!black}{Love Between the Raindrops}?\\[4pt]
\textbf{LLM output} &
\parbox[t]{0.92\linewidth}{\footnotesize\raggedright
Reasoning: \textbf{Step 1:} The director of the film \textit{Love Between the Raindrops} is Karel Kachy\v{n}a. \textbf{Step 2:} Karel Kachy\v{n}a was married to Alena Mihulov\'{a}. Conclusion: \textcolor{green!50!black}{Alena Mihulov\'{a}}.}\\[4pt]
\textbf{Judge / scores} &
\parbox[t]{0.92\linewidth}{%
\scoretag[7.2em]{green!18}{correct (1)} \quad
\scoretag{green!18}{S1: 1} \scoretag{green!18}{S2: 1}\\[4pt]
Alignment-aware UQ: 0.326 \quad Calibrated confidence: \scoretag{green!18}{0.997}}\\[5pt]
\textbf{Interpretation} &
This transfer example shows that the gain is not limited to longer chains. Even a short \textcolor{blue!70!black}{director $\rightarrow$ spouse} composition is initially scored conservatively by the alignment-aware lightweight UQ stage, but the reasoning consistency-aware calibrator identifies a fully supported relational bridge and upgrades confidence to near certainty.
\end{tabular}
\end{minipage}}}

\vspace{8pt}
\noindent\makebox[\textwidth][c]{\fcolorbox{red!65!black}{red!4}{
\begin{minipage}[t]{\dimexpr\appendixcardwidth-2\fboxsep-2\fboxrule\relax}
\small
\renewcommand{\arraystretch}{1.22}
\cardtitle{red!65!black}{Example F: OOD drift lowers confidence}
\begin{tabular}{@{}p{0.18\linewidth}p{0.76\linewidth}@{}}
\textbf{Question} &
What is the seat of the county sharing a border with the county where \textcolor{red!70!black}{Tenement} was founded?\\[4pt]
\textbf{LLM output} &
\parbox[t]{0.92\linewidth}{\footnotesize\raggedright
Reasoning: \textbf{Step 1:} Tenement was founded in the county of Baranya. \textbf{Step 2:} The county of Baranya shares a border with the county of Somogy. \textbf{Step 3:} The capital of Somogy county is Kaposv\'{a}r. Conclusion: \textcolor{red!70!black}{Kaposv\'{a}r}.}\\[4pt]
\textbf{Judge / scores} &
\parbox[t]{0.92\linewidth}{%
\scoretag[7.2em]{red!18}{incorrect (0)} \quad
\scoretag{red!18}{S1: 0} \scoretag{red!18}{S2: 0} \scoretag{red!18}{S3: 0}\\[4pt]
Alignment-aware UQ: 0.714 \quad Calibrated confidence: \scoretag{red!18}{0.375}}\\[5pt]
\textbf{Interpretation} &
This MuSiQue case highlights a different transfer failure from Example D. The chain looks geographically coherent on the surface, but it starts from an unsupported county assignment and then compounds that error through a neighboring-county hop. Reasoning consistency-aware calibration reacts to this \textcolor{red!70!black}{distribution drift} by cutting confidence substantially instead of trusting the fluent final conclusion.
\end{tabular}
\end{minipage}}}

\end{document}